\documentclass{article} %
\usepackage{colm2024_conference}

\usepackage{booktabs}
\usepackage{graphicx}
\usepackage{enumitem}
\usepackage{wrapfig}
\usepackage{algorithm}
\usepackage{algpseudocode}
\usepackage{natbib}
\usepackage{makecell}
\usepackage{bbm}
\usepackage{array}
\usepackage{amsmath} 
\usepackage{amssymb}
\usepackage{amsfonts}
\usepackage{multirow}
\usepackage{verbatim}
\usepackage{caption}
\usepackage{longtable}
\usepackage{supertabular}
\usepackage{hyperref}
\usepackage{CJKutf8}
\usepackage[utf8]{inputenc} %
\usepackage[T1]{fontenc} 
\usepackage[french,vietnamese,mongolian,greek,english]{babel}
\usepackage{afterpage}
\usepackage{tablefootnote}
\usepackage{xspace}
\usepackage{textcomp}
\usepackage{lscape} 
\usepackage{siunitx}
\usepackage{listings}
\usepackage{xcolor}
\usepackage{adjustbox}
\definecolor{dt}{gray}{0.7}
\definecolor{tongyi-purple}{RGB}{97,92,237}
\colorlet{tongyi-purple-alpha}{tongyi-purple!38}

\newcommand{\rev}[1]{#1}
\newcommand{\revdel}[1]{}
\newcommand{\pahfix}[1]{#1}
\newcommand{\pahadd}[1]{#1}

\usepackage{pifont}       %
\usepackage{bbding}       %
\usepackage{fontawesome}

\usepackage{scrextend}

\usepackage{tgpagella}
\usepackage{latexsym}
\usepackage{microtype}
\definecolor{mydarkblue}{rgb}{0,0.08,0.45}
\definecolor{citecolor}{HTML}{0071BC}
\usepackage{url}            %
\usepackage{nicefrac}       %
\usepackage{changepage}
\usepackage{xargs}          %
\usepackage{subcaption}
\usepackage{endnotes}

\usepackage{pgfplots}
\usetikzlibrary{pgfplots.groupplots}
\pgfplotsset{compat=1.3}
\usepackage{tikz}
\usetikzlibrary{patterns}

\usepackage[most]{tcolorbox}
\usepackage{fvextra}
\usepackage[capitalize,noabbrev]{cleveref}
\crefname{section}{Section}{Sections}
\Crefname{section}{Section}{Sections}
\crefname{table}{Table}{Tables}
\crefname{figure}{Figure}{Figures}
\crefname{algorithm}{Algorithm}{Algorithms}
\crefname{equation}{Eq.}{Eqs.}
\crefname{appendix}{Appendix}{Appendices}
\crefformat{section}{Section #2#1#3}
\usepackage{multicol}
\usepackage{fancyvrb} 
\newsavebox{\myverbsec}
\usepackage{titlesec}
\titleformat*{\section}{\large\bfseries}

\usepackage{nicematrix} %
\usepackage{arydshln}

\usepackage{dblfloatfix}

\makeatletter
\DeclareRobustCommand\onedot{\futurelet\@let@token\@onedot}
\def\@onedot{\ifx\@let@token.\else.\null\fi\xspace}

\title{Pistis Technical Report}

\author{
\bf Pistis Team, ByteDance}

\newcommand{\new}{\marginpar{NEW}}

\begin{document}
\maketitle
\vspace{-2.2em}

\begin{center}
\begin{tabular}{@{}c@{\hspace{0.55em}}l@{}}
\faGithub & \url{https://pististeam.github.io/} 
\end{tabular}
\end{center}

\begin{abstract}

We introduce the \textbf{Pistis} model family, comprising 27B- and 9B-parameter multimodal large language models built on Qwen3.6 and Qwen3.5, respectively, and developed through a general and scalable post-training framework. The framework first establishes a strong foundation through large-scale multimodal supervised fine-tuning (SFT). Building on this SFT foundation, we propose \textbf{Interleaved Distillation and Reinforcement Learning (IDRL)}, a novel post-training paradigm that tightly integrates on-policy distillation and reinforcement learning within a single training loop. By alternating between the two objectives, rather than optimizing either in isolation or combining them in a static joint loss, IDRL enables more effective knowledge transfer, greater optimization stability, and more precise credit assignment for long-horizon agentic trajectories, leading to stronger performance while mitigating common capability trade-offs.
At both model scales, the framework produces two specialized variants: \textbf{Pistis-Thinking}, designed to strengthen deep multimodal reasoning, and \textbf{Pistis-Agentic}, which additionally incorporates agentic trajectory data to support long-horizon planning, iterative reasoning, and tool use.
Pistis-Agentic is particularly strong in multimodal search. Both scales outperform their corresponding base models.
Beyond model-parameter optimization, we further introduce \textbf{Pistis-Auto-Harnessing (PAH)}, a system-level method that automatically improves the agent's inference harness through iterative optimization. Experiments demonstrate that PAH enhances the model performance without updating the model parameters or increasing the interaction budget. 
\end{abstract}

\section{Introduction}
\label{sec:intro}

\begin{figure*}[h]
\begin{center}
   \includegraphics[width=0.99\linewidth]{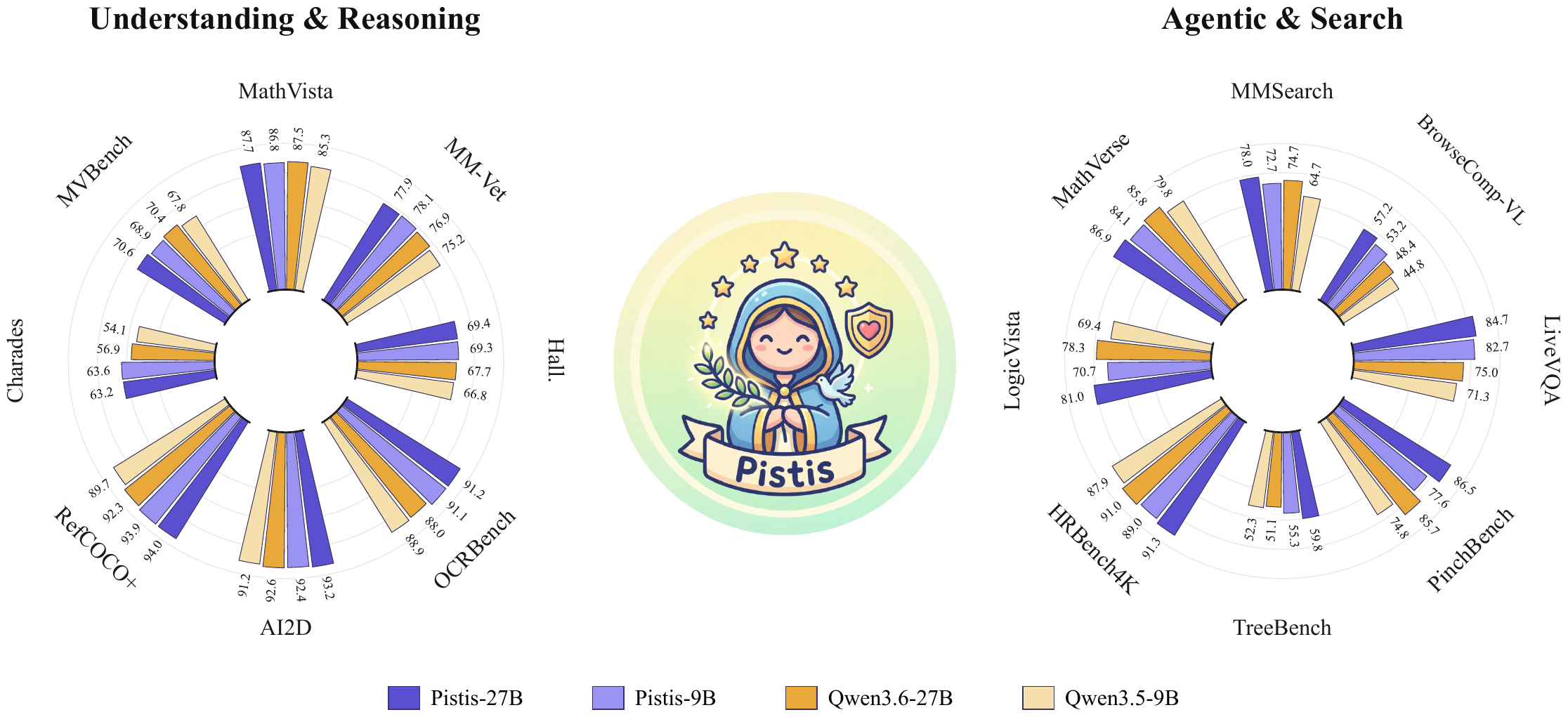}
\end{center}
   \caption{Overview of Pistis capabilities and performance.
The understanding and reasoning panel compares Pistis-Thinking with the corresponding base models on MathVista-mini, MM-Vet, HallusionBench, OCRBench, AI2D-test, RefCOCO+ testA, Charades-STA (64 frames), and MVBench (8 frames)~\citep{lu2023mathvista,yu2023mmvet,guan2024hallusionbench,liu2024ocrbench,kembhavi2016diagram,yu2016context,gao2017tall,li2024mvbench}. The agentic and search panel compares Pistis-Agentic with the corresponding base models on MMSearch, BrowseComp-VL, LiveVQA, PinchBench, TreeBench, HRBench4K, LogicVista, and MathVerse-mini~\citep{jiang2024mmsearch,geng2025webwatcher,fu2025livevqa,kilocode2026pinchbench,wang2025treebench,wang2024hrbench,xiao2024logicvista,zhang2024mathverse}. All values match Tables~\ref{tab:results_small} and~\ref{tab:results_agentic}.}

\label{fig:teaser}
\end{figure*}

Multimodal large language models (MLLMs) have made rapid progress in understanding images, videos, and documents, driven by strong foundation architectures such as \revdel{Qwen3-VL} \rev{Qwen3.5 and Qwen3.6}. Despite these advances, post-training remains a key bottleneck for unlocking higher-level reasoning, robust generalization, and real-world usability. Existing pipelines usually apply supervised fine-tuning and reinforcement learning in separate stages, leaving their complementary strengths underused: distillation provides dense supervision but can keep the student close to the teacher's distribution; reinforcement learning optimizes task reward but often collapses policy entropy and destabilizes training; and long-horizon agentic tasks require process supervision that a final-answer reward alone cannot provide.

In this work, we present the \textbf{Pistis} model family, comprising 27B- and 9B-parameter multimodal large language models built on Qwen3.6 and Qwen3.5, respectively, together with a general post-training framework. The framework first establishes a shared reasoning foundation through large-scale multimodal SFT on structured reasoning data. For the agentic variants, we augment this stage with agentic trajectories covering long-horizon planning, iterative reasoning, and tool use across textual and multimodal domains.

Building on this foundation, we propose \textbf{Interleaved Distillation and Reinforcement Learning (IDRL)}, a training paradigm that integrates on-policy distillation and reinforcement learning within a single training loop. Standard multimodal post-training treats distillation and RL as disjoint stages, preventing their signals from interacting during optimization. On-policy distillation provides dense token-level supervision by training the student to match the teacher's distribution on student-generated sequences~\citep{agarwal2024onpolicy,lu2025onpolicydistillation}, but it is sensitive to the initial student--teacher overlap and can fail without a suitable cold start~\citep{li2026rethinking}. IDRL retains both objectives but alternates between their updates rather than applying them sequentially or merging them into a single weighted loss. The two objectives can therefore shape each other across training phases without competing within the same update. Empirically, this interleaving improves exploration, stabilizes optimization, and outperforms pure on-policy distillation (OPD), pure RL, and their static combination in our ablations. For long-horizon agentic trajectories, IDRL also incorporates step-level positive-advantage suppression (PAS) to prevent rejected or ineffective intermediate actions from receiving positive credit solely because the final trajectory succeeds, thereby improving credit assignment. IDRL thus transfers knowledge from strong teachers while optimizing task-level objectives across diverse verifiable tasks, ranging from reasoning-centric problems to interactive agentic settings such as tool-integrated reasoning and deep research.

Complementing this model-level optimization, we introduce \textbf{Pistis-Auto-Harnessing (PAH)}, a system-level method that improves inference-time orchestration while keeping the agentic policy and tool interface fixed. An Optimization Agent uses development trajectories to propose and validate bounded harness revisions, retaining a candidate only when it improves a pre-specified development metric; the selected harness is then frozen for evaluation. We instantiate PAH on multimodal search, but its trace-guided propose--validate--update procedure applies more broadly whenever an executable harness and measurable development feedback are available.

Using this framework, we instantiate Thinking and Agentic variants at both the 27B and 9B scales. The Thinking models build on reasoning-data SFT to strengthen deep multimodal reasoning, whereas the Agentic models additionally incorporate agentic trajectories and are specialized for long-horizon, tool-integrated interaction. On the 24 non-grounding benchmarks, Pistis-27B-Thinking is comparable to Qwen3.8-27B (82.3 vs.\ 82.4), while it achieves the highest grounding average of 80.5, exceeding Qwen3.8-27B by 8.6 points. Pistis-9B-Thinking achieves the highest averages among the compared models at a similar scale on both benchmark groups. The Agentic variants also achieve higher overall averages than their corresponding base models, with particularly strong gains in multimodal search. Relative to Qwen3.6-27B, Pistis-27B-Agentic improves BrowseComp-VL, MMSearch, VDR-testmini, and LiveVQA by 8.8, 3.3, 3.2, and 9.7 points, respectively. The corresponding gains for Pistis-9B-Agentic over Qwen3.5-9B are 8.4, 8.0, 3.2, and 11.4 points (Figure~\ref{fig:teaser} and Table~\ref{tab:results_agentic}). Together, these results demonstrate strong generalization across model scales and task settings.

\noindent Our main contributions are:
\begin{itemize}[leftmargin=*,itemsep=2pt,topsep=2pt]
\item \textbf{Pistis model family.} We develop \textbf{Pistis-Thinking} models for deep multimodal reasoning and \textbf{Pistis-Agentic} models for long-horizon, tool-integrated interaction at both the 27B and 9B scales.
\item \textbf{Interleaved Distillation and Reinforcement Learning.} We propose IDRL, which alternates on-policy distillation and RL updates within one training loop, rather than running them as separate stages or combining them in a static joint loss. This design preserves policy entropy, reduces objective interference, and stabilizes optimization. For long-horizon agentic training, it incorporates PAS to improve credit assignment over intermediate actions.
\item \textbf{Pistis-Auto-Harnessing.} We propose an automated closed outer loop that iteratively improves the inference workflow, prompts, skills, evidence representation, and routing logic from development trajectories while keeping the Pistis-Agentic policy and tool interface fixed.
\item \textbf{Empirical validation.} Across public multimodal and agentic benchmarks, the 27B and 9B Pistis models achieve higher overall averages than their corresponding base models, with particular strengths in multimodal search and Claw-Style interaction. Training-dynamics and downstream ablations validate IDRL, including gains on the Claw-Style task group. PAH is further evaluated on a multimodal search benchmark, VDR-testmini, under the same environment-interaction limit.
\end{itemize}

\section{Approach}
\label{sec:approach}

\begin{figure*}[t]
    \centering
    \includegraphics[width=\textwidth]{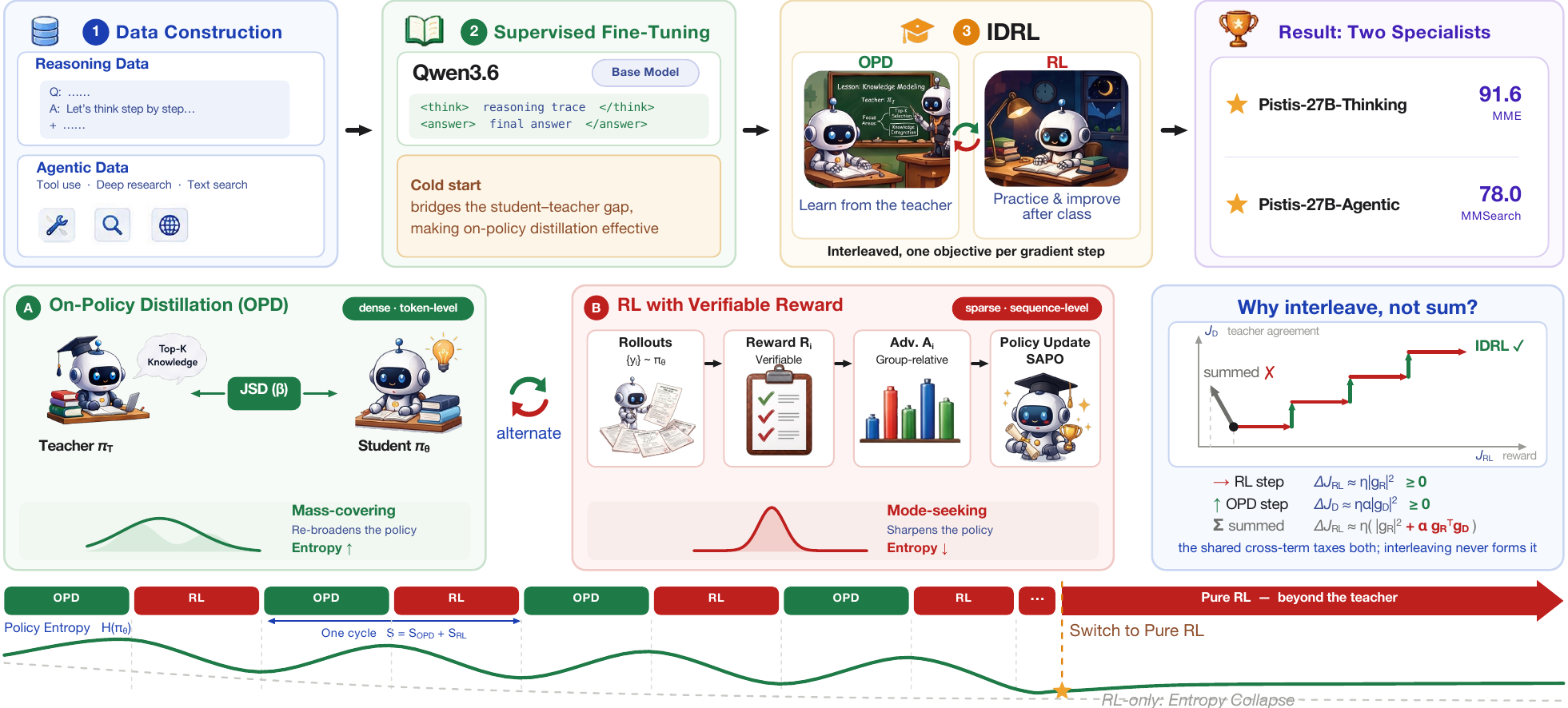}
    \caption{Overview of Interleaved Distillation and Reinforcement Learning (IDRL).}
    \label{fig:method_idrl}
\end{figure*}

\rev{The Pistis family is built on Qwen3.5-9B and Qwen3.6-27B base models and follows a shared post-training framework: large-scale multimodal supervised fine-tuning (SFT) followed by reinforcement-based optimization. The two scales share the same framework while using task-specific data compositions and optimization configurations. We first build a strong multimodal foundation from structured reasoning and agentic trajectories and then introduce Interleaved Distillation and Reinforcement Learning (IDRL) as our core algorithmic contribution. At inference time, we further introduce our automated harness optimization method---Pistis-Auto-Harnessing (PAH). Pistis-Thinking is trained with SFT on reasoning data and subsequently optimized on reasoning-centric tasks, whereas Pistis-Agentic additionally incorporates agentic trajectories during SFT and is specialized on interactive agentic tasks. Finally, we describe the infrastructure supporting efficient training, evaluation, and inference.}

\subsection{Supervised Fine-Tuning}
We construct a large-scale, diverse multimodal SFT corpus of approximately 3.2M QA pairs, comprising \textbf{reasoning data} and \textbf{agentic trajectory data}, together with a quality-control pipeline that filters noisy samples and strengthens the resulting SFT models. The reasoning data forms the shared SFT foundation for both models, while the agentic trajectory data is additionally included when fine-tuning \revdel{Pistis-8B-Agentic} \rev{Pistis-Agentic}, allowing it to build agentic skills on top of the same reasoning foundation.

\noindent \textbf{Reasoning Data.}
The reasoning data spans a broad range of domains and tasks, including mathematical reasoning; chart, figure, table, and document understanding; scientific and diagram reasoning; medical visual question answering; code generation; general logical reasoning; and general real-world visual question answering (including knowledge-based and scene-text reading). The data are aggregated from public repositories and carefully designed synthetic prompts, with high-quality responses generated by a strong teacher model under controlled prompting that elicits detailed step-by-step reasoning, grounded visual analysis, and consistent answer formulation. Each instance adopts a structured output format in which the intermediate reasoning trace is enclosed by \texttt{<think>}\dots\texttt{</think>}, followed by two newline characters and the final answer without \texttt{<answer>} tags; this separation enables reliable parsing and fine-grained supervision over both reasoning quality and answer correctness. All samples are normalized into a unified schema and pass a quality-control pipeline: we validate tag usage, logical completeness, and a minimum reasoning length, remove malformed or underspecified samples, and, for tasks with reference answers, perform answer-level consistency checks to discard mismatched or unverifiable outputs.

\revdel{The agentic corpus covers tool-integrated reasoning, multimodal deep research, and text-only search.}

\rev{\noindent \textbf{Agentic Trajectory Data.}
For the agentic SFT mixture, tool-integrated reasoning (TIR), search, and general agent trajectories account for approximately 40\%, 20\%, and 40\%, respectively. TIR covers mathematical reasoning, visual perception, and visual-logic tasks. Search data includes both text-based and multimodal retrieval. The remaining trajectories are high-quality open-source examples spanning software engineering, code generation, tool use, and long-horizon environment interaction.}

\rev{For self-generated TIR and search data, we sample up to four response trajectories for each query and evaluate them with task-specific rewards. We discard queries for which all sampled trajectories fail or all succeed. For a query with an intermediate success rate, we randomly retain one successful trajectory for SFT. This selection focuses supervision on moderately difficult examples, for which a successful solution is informative while substantial room for improvement remains. We further filter trajectories by final-answer correctness and replace correct but low-quality open-source reasoning with cleaner regenerated trajectories when appropriate.}

\noindent \textbf{Training Details.}
We train with the AdamW optimizer using a base learning rate of $1\times10^{-5}$ and a cosine decay schedule.
To improve efficiency and reduce memory fragmentation, we adopt the Liger kernel, apply sequence packing up to a maximum length of 32{,}768 tokens, and preserve native-resolution inputs for fine-grained visual perception. Training runs for three epochs with a global batch size of 1{,}536.
Supervised fine-tuning takes approximately 9{,}216 GPU-hours on the combined reasoning and agentic trajectory data.

\subsection{Interleaved Distillation and Reinforcement Learning}
Modern large language models increasingly rely on post-training to improve general capabilities and task performance. However, existing paradigms often create limited synergy between supervised distillation and reinforcement learning, leaving their complementary strengths under-exploited.

To address this limitation, we propose Interleaved Distillation and Reinforcement Learning (IDRL), a training paradigm that integrates distillation and RL through an interleaved optimization process. Specifically, IDRL alternates between On-Policy Distillation (OPD) and RL during training, with flexible step budgets for each phase. This design enables dynamic interaction between the two learning signals, rather than treating them as isolated or sequential stages.

Intuitively, RL and OPD oppose each other: RL sharpens the policy toward high-reward outputs, reducing entropy, whereas OPD draws the student toward the teacher's broader distribution with a dense token-level signal that preserves entropy. Summing the objectives forces these opposing updates into a single gradient, where they compete directly, whereas alternating them lets each step follow a single objective while the two still influence each other across phases. We formalize this schedule next, and analyze its effect on entropy and stability in the following subsections.

Formally, let $s$ denote the current global training step. We define the durations, or step budgets, for the OPD and RL phases as $S_{\text{OPD}}$ and $S_{\text{RL}}$, respectively. The optimization strategy $\Pi_s$ at step $s$ is governed by a periodic scheduling function:

\begin{equation}
\Pi_s = 
\begin{cases} 
\text{OPD}, & \text{if } (s \bmod S) < S_{\text{OPD}} \text{ and } s < S^* \\
\text{RL}, & \text{otherwise}
\end{cases}
\end{equation}

where $S = S_{\text{OPD}} + S_{\text{RL}}$ represents the total length of one full interleaved cycle. We switch to pure RL after $S^*$ steps, where $S^*$ is determined by monitoring the entropy plateau. 
The transition to pure RL is motivated by two observations: (i) once the entropy plateaus, continued OPD contributes little additional output diversity, and (ii) by continually regularizing the student toward the teacher's distribution, it upper-bounds the student at the teacher's capability, foreclosing any super-teacher gains.
We apply IDRL to the 9B variants, using reasoning-centric mixtures for Pistis-9B-Thinking and interactive agentic mixtures for Pistis-9B-Agentic. The corresponding 27B variants are optimized with pure RL under the same RL hyperparameters and serve as the frozen OPD teachers for 9B models.
For reproducibility, Table~\ref{tab:idrl_training_config} summarizes the training paradigm, teacher assignment, and optimization hyperparameters for each Pistis variant.
\begin{table*}[t]
\centering
\caption{Post-training configurations and compute. The 9B variants use IDRL, whereas the 27B variants use pure RL. $S_{\mathrm{OPD}}$ and $S_{\mathrm{RL}}$ denote the numbers of OPD and RL steps in each interleaved cycle, respectively, and $S^*$ denotes the step at which IDRL switches to pure RL.}
\label{tab:idrl_training_config}
\renewcommand{\arraystretch}{1.15}
\begin{adjustbox}{max width=\textwidth}
\begin{tabular}{lcccc}
\toprule
\textbf{Configuration} &
\shortstack{\textbf{Pistis-9B} \\ \textbf{Thinking}} &
\shortstack{\textbf{Pistis-27B} \\ \textbf{Thinking}} &
\shortstack{\textbf{Pistis-9B} \\ \textbf{Agentic}} &
\shortstack{\textbf{Pistis-27B} \\ \textbf{Agentic}} \\
\midrule
Training paradigm                       & IDRL & Pure RL & IDRL & Pure RL \\
OPD teacher                              & Pistis-27B-Thinking & -- & Pistis-27B-Agentic & -- \\
$S_{\mathrm{OPD}}$ (steps/cycle)       & 5 & -- & 5& -- \\
$S_{\mathrm{RL}}$ (steps/cycle)        & 5 &-- & 5&-- \\
Switch step $S^*$                       & 300 & -- & 100& -- \\
Distillation weight $\alpha$            & 1 & -- & 1& -- \\
JSD coefficient $\beta$                 & 0.5 & -- & 0.5& -- \\
Teacher top-$k$                          & 50 & -- & 50 & -- \\
Maximum responses length                & 32k& 32k & 100k & 100k\\
SAPO temperature $\tau_{\mathrm{pos}}$ & 1 &1 & 1& 1\\
SAPO temperature $\tau_{\mathrm{neg}}$ & 1.05 &1.05 &1.05 &1.05 \\
Learning rate                           & 1e-6 & 1e-6 & 1e-6 & 1e-6 \\
\bottomrule
\end{tabular}
\end{adjustbox}
\end{table*}

In the following, we detail the two components of IDRL, on-policy distillation and reinforcement learning, and then analyze why interleaving them preserves entropy and stabilizes optimization.

\subsubsection{On-Policy Distillation}
We use on-policy distillation to transfer teacher behavior to the student under the student's own sampled contexts. We formulate this distillation process following the Generalized Knowledge Distillation (GKD) framework of \citet{agarwal2024onpolicy}, instantiated with the Jensen--Shannon divergence (JSD) as the training objective.

\paragraph{Setup.}
Let $\pi_\theta$ denote the student policy being optimized, and $\pi_T$ the teacher policy corresponding to a prompt $x$ drawn from distribution $\mathcal{D}$. Given an output sequence $y$ sampled on-policy from the student, the token-level discrepancy between teacher and student is measured by the generalized JSD at each position $t$:
\begin{equation}
\label{eq:jsd}
\ell^{(t)}_{\text{OPD}}= D_{\mathrm{JSD}(\beta)}\!\left(\pi_T(\cdot\mid x,y_{<t}) \,\|\, \pi_\theta(\cdot\mid x,y_{<t})\right)
= \beta\, D_{\mathrm{KL}}\!\left(\pi_T \,\big\|\, M\right)
+ (1-\beta)\, D_{\mathrm{KL}}\!\left(\pi_\theta \,\big\|\, M\right),
\end{equation}
where $M = \beta\,\pi_T(\cdot\mid x,y_{<t}) + (1-\beta)\,\pi_\theta(\cdot\mid x,y_{<t})$ is the mixture distribution and $\beta \in (0,1)$ controls the asymmetry between the teacher-to-mixture and student-to-mixture KL terms. Compared to the plain reverse KL objective, the JSD formulation is bounded and provides gradient signal from both directions, which we find leads to more stable training in practice.

\paragraph{Top-$k$ Vocabulary Restriction.}
Computing the full-vocabulary divergence at every token is expensive and dominated by near-zero probability mass. We therefore restrict the computation to the top-$k$ tokens selected by the \emph{teacher} distribution at each decoding step. Let $\mathcal{V}_t \subseteq \mathcal{V}$ denote this top-$k$ index set with $|\mathcal{V}_t| = k$. We approximate $\pi_T(v \mid x, y_{<t}) \approx 0$ for all $v \notin \mathcal{V}_t$. In practice, this approximation is acceptable for current LLMs, whose output distributions are typically sharp and concentrated on a small fraction of the vocabulary.

Under this approximation, we expand each KL term in the JSD separately over $\mathcal{V}_t$ and its complement $\mathcal{V} \setminus \mathcal{V}_t$. For the first term in Eq.~\ref{eq:jsd}, since $\pi_T(v\mid x,y_{<t}) \approx 0$ for $v \notin \mathcal{V}_t$, the sum reduces directly to the top-$k$ tokens:
\begin{equation}
D_{\mathrm{KL}}\!\left(\pi_T \,\big\|\, M\right)
= \sum_{v \in \mathcal{V}_t} \pi_T(v\mid x,y_{<t})\log\frac{\pi_T(v\mid x,y_{<t})}{M(v)}
+ \underbrace{\sum_{v \notin \mathcal{V}_t} \pi_T(v\mid x,y_{<t})\log\frac{\pi_T(v\mid x,y_{<t})}{M(v)}}_{\approx\;0}.
\end{equation}
For the second term, we first note that outside $\mathcal{V}_t$, the mixture simplifies to $M(v) = \beta\,\pi_T(v\mid x,y_{<t}) + (1-\beta)\,\pi_\theta(v\mid x,y_{<t}) \approx (1-\beta)\,\pi_\theta(v\mid x,y_{<t})$. Substituting this approximation into $D_{\mathrm{KL}}\!\left(\pi_\theta \,\big\|\, M\right)$:
\begin{align}
D_{\mathrm{KL}}\!\left(\pi_\theta \,\big\|\, M\right)
&= \sum_{v \in \mathcal{V}_t} \pi_\theta(v\mid x,y_{<t})\log\frac{\pi_\theta(v\mid x,y_{<t})}{M(v)}
+ \sum_{v \notin \mathcal{V}_t} \pi_\theta(v\mid x,y_{<t})\log\frac{\pi_\theta(v\mid x,y_{<t})}{M(v)} \nonumber\\
&\approx \sum_{v \in \mathcal{V}_t} \pi_\theta(v\mid x,y_{<t})\log\frac{\pi_\theta(v\mid x,y_{<t})}{M(v)}
+ \log\frac{1}{1-\beta} \sum_{v \notin \mathcal{V}_t} \pi_\theta(v\mid x,y_{<t}) \nonumber\\
&= \sum_{v \in \mathcal{V}_t} \pi_\theta(v\mid x,y_{<t})\log\frac{\pi_\theta(v\mid x,y_{<t})}{M(v)}
+ \log\frac{1}{1-\beta} \left(1-\sum_{v \in \mathcal{V}_t} \pi_\theta(v\mid x,y_{<t})\right)
\end{align}
Combining both terms, and omitting the shared conditioning on $(x,y_{<t})$ for readability, the per-token OPD objective $\ell^{(t)}_{\text{OPD}}$ is:

\begin{equation}
\ell^{(t)}_{\text{OPD}} = \beta \sum_{v\in\mathcal{V}_t} \pi_T(v)\log\frac{\pi_T(v)}{M(v)} + (1-\beta) \left[ \sum_{v\in\mathcal{V}_t} \pi_\theta(v) \log\frac{\pi_\theta(v)}{M(v)} + \log\frac{1}{1-\beta} \left(1-\sum_{v\in\mathcal{V}_t}  \pi_\theta(v) \right) \right]
\end{equation}
Note that $\pi_\theta$ is \emph{not} renormalized over $\mathcal{V}_t$. The residual term $1-\sum_{v\in\mathcal{V}_t}  \pi_\theta(v)$ instead analytically accounts for the student mass outside the teacher's top-$k$ support, avoiding the distortion introduced by renormalization while remaining computationally efficient.

\paragraph{Choice of divergence and top-$k$.}
The choice of divergence is what makes OPD an effective entropy regularizer inside IDRL. Minimizing a reverse KL $D_{\mathrm{KL}}(\pi_\theta\,\|\,\pi_T)$ is \emph{mode-seeking}: it is zero-forcing and drives the student to concentrate probability on a few teacher modes, which, like reward-driven RL, reduces policy entropy. A forward KL $D_{\mathrm{KL}}(\pi_T\,\|\,\pi_\theta)$ is instead \emph{mass-covering}, forcing the student to spread probability across the teacher's support and thereby preserving entropy. The generalized JSD interpolates between these two regimes, and the top-$k$ set controls how much of the teacher's support the student must cover. With a sufficiently broad top-$k$, OPD pulls the student toward the teacher's comparatively high-entropy distribution and counteracts the entropy collapse induced by RL. This mechanism matches the OPD-variant ablation in Section~\ref{sec:exp} (Figure~\ref{fig:OPD_analysis}): the mode-seeking reverse KL collapses entropy, the overly narrow top-$5$ JSD still collapses because coverage is insufficient, whereas the broader top-$50$ JSD sustains high and stable entropy. We therefore set $k=50$. Crucially, this top-$k$ objective remains informative only when the student already places appreciable mass on the teacher's top-$k$ tokens, i.e., when the two distributions overlap; we find this condition breaks down at initialization in multimodal settings, which we address next.

\paragraph{Applying OPD to Multimodal Settings.}

\begin{figure}[t]
  \centering
  \begin{minipage}[c]{0.5\linewidth}
    \centering
    \captionof{table}{Geo3K accuracy under different training configurations.}
    \label{tab:multimodal_opd}
    \begin{tabular}{lc}
      \toprule
      Model & Geo3K \\
      \midrule
      Qwen3.6-27B (Teacher) &  70.2 \\
      Qwen3.5-9B (Student) & 65.0 \\
      \quad + off-policy SFT & 65.4 \\
      \quad + on-policy OPD & 65.7 \\
      \bottomrule
    \end{tabular}
  \end{minipage}
  \hfill
  \begin{minipage}[c]{0.46\linewidth}
    \centering
    \includegraphics[width=\linewidth]{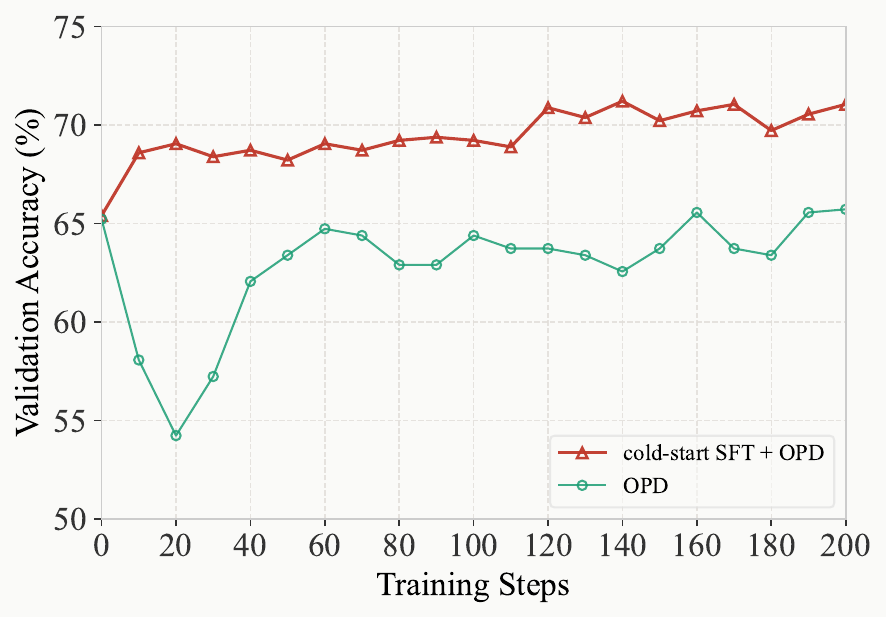}
    \captionof{figure}{Geo3K accuracy over training steps with different training strategies.}
    \label{fig:geo3k_opd}
  \end{minipage}
\end{figure}

Directly applying OPD to MLLMs yields only limited gains. As shown in Table~\ref{tab:multimodal_opd}, with a Qwen3.5-9B student and a Qwen3.6-27B teacher on Geo3K, off-policy SFT and on-policy OPD improve the base student only marginally, from 65.0 to 65.4 and 65.7, respectively. The modest gain from OPD suggests that on-policy sampling alone does not resolve the difficulty. Instead, the objective may be limited by a mismatch between the initial student and teacher distributions.

To investigate, we prepend a cold-start SFT phase before OPD, fine-tuning the student on teacher-generated responses before switching to on-policy OPD. As shown in Figure~\ref{fig:geo3k_opd}, direct OPD drops sharply at the start, whereas cold-start OPD remains stable, improves consistently, and converges above the base student. This result suggests that \textbf{the initial distribution gap between student and teacher is the key obstacle, and that bridging it via cold-start SFT is sufficient to restore effective distillation}. While \citet{li2026rethinking} identify low initial token-distribution overlap as the governing failure condition for OPD in text-only settings, our results show that the same mechanism extends to multimodal models, and that cold-start SFT remains an effective remedy in this more challenging regime.

\subsubsection{Reinforcement Learning with Verifiable Reward}
\revdel{Pistis-8B-Thinking and Pistis-8B-Agentic are trained on separate task mixtures.}
\rev{We conduct RL over a broad spectrum of textual and multimodal tasks whose outputs can be scored reliably, in most cases by predefined rules or executable programs. Pistis-Thinking is trained on reasoning and grounding tasks, spanning STEM and visual reasoning, image grounding, visual counting, and temporal video grounding. For Pistis-Agentic, IDRL focuses on two core capabilities: TIR and search, which account for approximately 60\% and 40\% of the mixture, respectively. The TIR portion covers mathematical reasoning, visual perception, and visual-logic reasoning; the search portion covers general web search, visual fact retrieval, and visual document retrieval. The training corpus is assembled from open-source and proprietary resources with strict preprocessing and human annotation.}

\revdel{Qwen3-VL-235B-A22B is used as a trajectory generator.}
\rev{For the largest task family, visual reasoning, we initially curate roughly 80K candidate STEM problems from open-source platforms and proprietary K-12 data, using an LLM to remove proof-style questions, convert multiple-choice items into open-ended form (reducing reward hacking), filter by model-estimated difficulty, and exclude problems solvable without the image. For multimodal queries, we sample 16 candidate responses per query from strong MLLMs and discard queries whose responses are all incorrect. After source-specific filtering and deduplication, this pool is combined with the other task sources; pilot RL experiments then prune sources with low improvement potential, yielding a final RL corpus of roughly 30K high-quality queries. During training, we again sample 16 responses per query, remove overly easy queries (pass rate above 90\%), and merge task-specific data into mixed-task batches with a fixed, empirically tuned sampling ratio; grounding data combines general-purpose and GUI tasks.}

\paragraph{Optimization with SAPO.}
For policy optimization we adopt SAPO~\citep{gao2025sapo}, a smooth, adaptive policy-gradient algorithm. For each query $x$ we sample a group of $G$ responses $\{y_i\}_{i=1}^{G}$ from $\pi_{\theta_{\text{old}}}$, score each with a verifiable reward $R_i$, and assign a group-relative advantage shared across the tokens of a response,
\begin{equation}
A_i = \frac{R_i - \mathrm{mean}(\{R_j\}_{j=1}^{G})}{\mathrm{std}(\{R_j\}_{j=1}^{G})}.
\end{equation}
SAPO then maximizes
\begin{equation}
\mathcal{J}_{\text{RL}}(\theta) = \mathbb{E}_{x\sim\mathcal{D},\,\{y_i\}\sim\pi_{\theta_{\text{old}}}}\!\left[\frac{1}{G}\sum_{i=1}^{G}\frac{1}{|y_i|}\sum_{t=1}^{|y_i|} f_{i,t}\!\big(r_{i,t}(\theta)\big)\,A_i\right],\qquad r_{i,t}(\theta)=\frac{\pi_\theta(y_{i,t}\mid x,y_{i,<t})}{\pi_{\theta_{\text{old}}}(y_{i,t}\mid x,y_{i,<t})},
\end{equation}
where, in place of the hard PPO-style clipping used by GRPO, the importance ratio is reweighted by a smooth, temperature-controlled gate
\begin{equation}
f_{i,t}(r) = \frac{4}{\tau_{i,t}}\,\sigma\!\big(\tau_{i,t}(r-1)\big),\qquad
\tau_{i,t} = \begin{cases}\tau_{\text{pos}}, & A_i > 0,\\[2pt] \tau_{\text{neg}}, & A_i \le 0,\end{cases}
\end{equation}
with $\sigma$ the sigmoid and asymmetric temperatures $\tau_{\text{pos}},\tau_{\text{neg}}$. The induced update weight $f_{i,t}'(r)=4\,\sigma(\tau_{i,t}(r-1))\,(1-\sigma(\tau_{i,t}(r-1)))$ peaks at $r_{i,t}=1$ (on-policy) and decays smoothly as the ratio departs from $1$, so off-policy updates are attenuated continuously rather than clipped discontinuously, forming a smooth trust region that stabilizes training across task types and model scales.

\rev{\paragraph{Step-Level Positive-Advantage Suppression.}
Trajectory-level rewards alone can incorrectly reinforce intermediate tool calls in a successful trajectory. To improve credit assignment, we identify assistant steps that are rejected or do not contribute to later trajectory states, including malformed tool calls, repeated or substantially similar calls, calls issued after a final-answer instruction, and calls whose observations cannot be incorporated into the context. After advantage estimation, we set the positive advantages of all tokens in these steps to zero, while retaining zero or negative advantages. Thus, an ineffective intermediate action is not positively reinforced even if the trajectory eventually receives a positive outcome, but it remains penalized in unsuccessful trajectories. This asymmetric treatment separates the quality of intermediate actions from the final trajectory-level outcome and yields more precise credit assignment for multi-turn agent interaction.}

\paragraph{Reward System}
We supervise RL with a hybrid, task-aware reward formulation. Our guiding principle is to prefer rewards that can be \emph{automatically verified} by a rule or an executable check whenever the task admits one; such rewards are precise and reproducible, and they avoid the dominant failure mode of model-based judging, where the policy learns to exploit the judge rather than solve the task. Concretely, a global format reward first checks that the reasoning tags are present and correctly paired. Tasks with a single correct answer, including STEM problems, chart numerical questions, and OCR, receive a binary correctness reward through rule-based exact match, while spatial and temporal grounding receive a continuous IoU reward against the ground truth. For open-ended tasks where no deterministic check exists, namely long-document QA, general VQA, and agentic search, we fall back to model-based evaluation with a strong judge model. Table~\ref{tab:reward_system} summarizes the design for each task type.

\begin{table}[t]
\centering
\caption{Task-aware reward design in the IDRL stage of Pistis, including a global format constraint and domain-specific correctness rewards.}
\label{tab:reward_system}
\resizebox{\linewidth}{!}{%
\renewcommand{\arraystretch}{1.3} %
\begin{tabular}{@{} l l c c c p{0.55\linewidth} @{}}
\toprule
Reward scope & Task domain & Rule & Model & Binary & \multicolumn{1}{c}{Reward design details} \\
\midrule
Format & All Domains & \checkmark & & \checkmark & Score $1$ if \texttt{<think>} and \texttt{</think>} tags are present and strictly paired; otherwise $0$. \\
\midrule
\multirow{3}{*}{STEM}
 & Math & \checkmark & & \checkmark & \multirow{3}{=}{Numerical calculation \& multiple-choice: Rule-based exact match; score $1$ for correct, $0$ for incorrect.} \\
 & Physics & \checkmark & & \checkmark & \\
 & Chemistry & \checkmark & & \checkmark & \\
\midrule
\multirow{3}{*}{\begin{tabular}[c]{@{}l@{}}Long Document \\ Chart \& OCR\end{tabular}}
 & Long Document & & \checkmark & & Open-ended QA: Model-based evaluation using a strong judge model. \\
 & Chart & \checkmark & & \checkmark & Numerical calculation \& multiple-choice: Rule-based exact match; score $1$ for correct, $0$ for incorrect. \\
 & OCR & \checkmark & & \checkmark & Rule-based exact match; score $1$ for correct, $0$ for incorrect. \\
\midrule
General VQA & VQA & & \checkmark & & Open-ended QA: Model-based evaluation using a strong judge model. \\
\midrule
\multirow{2}{*}{Grounding}
 & Spatial & \checkmark & & & \multirow{2}{=}{Score equals the IoU between the prediction and ground truth.} \\
 & Temporal & \checkmark & & & \\
 \midrule
Agent & Search & & \checkmark & &  Model-based evaluation using a strong judge model. \\
\bottomrule
\end{tabular}%
}
\end{table}

\subsubsection{Benefits of Interleaved Distillation and Reinforcement Learning}
To understand why \emph{interleaving} the two objectives is more effective than simply combining them, we examine the objective optimized at each step. At step $s$, the schedule $\Pi_s$ defined above activates exactly one objective, so the per-step objective can be written as the single indicator-gated sum
\begin{equation}
\begin{aligned}
\mathcal{L}_{s}(\theta)
= -\,\mathbbm{1}[\Pi_s=\text{RL}]\,\mathcal{J}_{\text{RL}}(\theta)
+ \alpha\,\mathbbm{1}[\Pi_s=\text{OPD}]\,\mathcal{L}_{\text{OPD}}(\theta),
\end{aligned}
\end{equation}
where $\mathbbm{1}[\cdot]$ is the indicator function and $\alpha>0$ weights distillation. Here $\mathcal{J}_{\text{RL}}(\theta)$ is the reward-maximizing SAPO objective defined above, and $\mathcal{L}_{\text{OPD}}(\theta)=\mathbb{E}_{x\sim\mathcal{D},\,y\sim\pi_\theta}\big[\tfrac{1}{|y|}\sum_t \ell^{(t)}_{\text{OPD}}\big]$ is the OPD loss averaging the per-token JSD $\ell^{(t)}_{\text{OPD}}$ of Eq.~\ref{eq:jsd} (mixture $M=\beta\pi_T+(1-\beta)\pi_\theta$, all distributions conditioned on $(x,y_{<t})$). Since exactly one indicator is nonzero at any step, IDRL never optimizes the two terms at once, in contrast to the joint objective $\mathcal{L}_{\text{joint}}(\theta)=-\mathcal{J}_{\text{RL}}(\theta)+\alpha\mathcal{L}_{\text{OPD}}(\theta)$, which always sums them. The distillation phases act as a teacher anchor loosely analogous to the KL-to-reference term in KL-regularized RL, but stronger: the reference is a more capable teacher rather than a frozen copy of the policy, so distillation both transfers new capability and re-broadens the output distribution. Two properties explain why this alternation is preferable.

\noindent\textbf{(i) Distillation preserves entropy and provides a dense signal.}
The OPD phases minimize $\mathcal{L}_{\text{OPD}}$, which vanishes only when $\pi_\theta=\pi_T$ and thus pulls the student toward the teacher distribution. Through the identity $D_{\mathrm{JSD}(\beta)}(\pi_T\|\pi_\theta)=H(M)-\beta H(\pi_T)-(1-\beta)H(\pi_\theta)$, where $H(\cdot)$ denotes entropy and the teacher is fixed, the JSD objective includes an entropy-related term while also depending on the mixture entropy $H(M)$. Rather than relying on this term alone, the key effect is that the mass-covering component of JSD encourages the student to cover a broader teacher support than mode-seeking reverse KL, which empirically sustains higher policy entropy in our ablations. This teacher-anchored signal counteracts the entropy collapse driven by the reward term, keeping the policy's outputs diverse and complementing the mode-seeking-versus-mass-covering view from the on-policy distillation discussion above. The distillation signal is also \emph{dense}: it supplies a token-level target at every position, unlike the sparse, sequence-level scalar reward of RL, which improves credit assignment and stabilizes learning.

\noindent\textbf{(ii) Interleaving avoids gradient conflict.}
Writing the two updates explicitly shows when summing them is harmful. Consider a single decoding position with policy distribution $\pi_\theta(\cdot\mid x,y_{<t})$ and teacher $\pi_T(\cdot\mid x,y_{<t})$; both gradients are combinations of the token score functions $\nabla_\theta\log\pi_\theta(v\mid x,y_{<t})$. At the on-policy point the SAPO gate satisfies $f_{i,t}'(1)=1$, so the RL gradient reduces to the policy gradient
\begin{equation}
g_R = \nabla_\theta\mathcal{J}_{\text{RL}}(\theta)\big|_{\theta=\theta_{\text{old}}} = \mathbb{E}_{x\sim\mathcal{D},\,\{y_i\}\sim\pi_{\theta_{\text{old}}}}\!\Big[\tfrac{1}{G}\sum_{i=1}^{G}\tfrac{1}{|y_i|}\sum_t A_i\,\nabla_\theta\log\pi_\theta(y_{i,t}\mid x,y_{i,<t})\Big],
\end{equation}
which for $A_i>0$ raises the probability of the realized token $a=y_{i,t}$ and thus \emph{concentrates} mass (mode-seeking). Using $\partial D_{\mathrm{JSD}(\beta)}(\pi_T\|\pi_\theta)/\partial\pi_\theta(v)=(1-\beta)\log\frac{\pi_\theta(v)}{M(v)}$, the distillation ascent direction, holding the on-policy samples fixed as in on-policy distillation, is
\begin{equation}
g_D = -\nabla_\theta\mathcal{L}_{\text{OPD}}(\theta)\big|_{\theta=\theta_{\text{old}}} = -(1-\beta)\,\mathbb{E}_{x\sim\mathcal{D},\,y\sim\pi_{\theta_{\text{old}}}}\Big[\textstyle\tfrac{1}{|y|}\sum_t\sum_v \pi_\theta(v)\log\tfrac{\pi_\theta(v)}{M(v)}\,\nabla_\theta\log\pi_\theta(v)\Big],
\end{equation}
which raises probability on tokens with $\pi_\theta(v)<M(v)$ and thus \emph{spreads} mass toward the teacher mixture (mass-covering). The two act oppositely on the realized token $a$: up to positive normalization, the RL contribution to the coefficient of $\nabla_\theta\log\pi_\theta(a)$ is proportional to $A_i>0$, whereas the OPD contribution is proportional to $-(1-\beta)\,\pi_\theta(a)\log\frac{\pi_\theta(a)}{M(a)}$. Whenever the policy is already more confident on $a$ than the teacher, $\pi_\theta(a)>\pi_T(a)$ (equivalently $\pi_\theta(a)>M(a)$), this OPD contribution is negative, so the two contributions have opposite signs; when such over-confident tokens dominate the batch, the aggregated gradients conflict, $g_R^{\top}g_D<0$, precisely in the over-sharpened regime that OPD is meant to correct. The cost of this conflict is clearest in the first-order improvement each update produces. Summing the two signals, $\theta^{+}=\theta+\eta(g_R+\alpha g_D)$ with step size $\eta$, moves the two objectives by
\begin{equation}
\Delta\mathcal{J}_{\text{RL}}\approx\eta\big(\|g_R\|^{2}+\alpha\,g_R^{\top}g_D\big),
\qquad
\Delta\mathcal{J}_{\text{D}}\approx\eta\big(\alpha\|g_D\|^{2}+g_R^{\top}g_D\big),
\end{equation}
where $\mathcal{J}_{\text{D}}:=-\mathcal{L}_{\text{OPD}}$ is the distillation objective. The shared cross-term $g_R^{\top}g_D$ penalizes both updates: under conflict each objective improves less than it would in isolation, and once $g_R^{\top}g_D<-\|g_R\|^{2}/\alpha$ the summed step \emph{lowers} the reward objective even though it descends the joint loss. Interleaving never forms this cross-term, because each step carries a single gradient: an RL step gives $\Delta\mathcal{J}_{\text{RL}}\approx\eta\|g_R\|^{2}\ge0$ and an OPD step gives $\Delta\mathcal{J}_{\text{D}}\approx\eta\alpha\|g_D\|^{2}\ge0$, each non-negative to first order regardless of the angle between $g_R$ and $g_D$, and vanishing only at a stationary point of the active objective. Every interleaved update is therefore an uncorrupted, single-objective step, whereas the summed update can stall or reverse one of the two. This per-step guarantee is what keeps optimization well-behaved; the complementary, cross-phase effect, in which OPD periodically restores the entropy and exploration that the multi-step RL phases then exploit, is an empirical property, which we examine next.

These properties are borne out empirically (Section~\ref{sec:exp}): OPD raises and sustains policy entropy where pure RL collapses it, and the interleaved variant shows smoother gradient norms and a clean alternating entropy pattern while the summed (RL+OPD) variant is noisier (Figure~\ref{fig:IDRL_analysis}). Once the student's entropy plateaus, switching entirely to RL after $S^*$ steps leverages the enlarged exploration space for stronger and more stable improvement.

\subsection{\texorpdfstring{\pahfix{Closed-Loop} Auto-Harnessing for Multimodal Search}{Closed-Loop Auto-Harnessing for Multimodal Search}}
\label{sec:auto_harness}

\begin{figure*}[t]
\centering
\begin{adjustbox}{max width=\textwidth}
\begin{tikzpicture}[
  font=\scriptsize,
  panel/.style={draw=#1!60, fill=#1!5, rounded corners=5pt, line width=0.9pt},
  stage/.style={draw=#1!75, fill=white, rounded corners=3pt, align=center,
                minimum height=0.95cm, text width=2.0cm, inner sep=3pt, line width=0.7pt},
  card/.style={draw=#1!75, fill=#1!8, rounded corners=3pt, align=center,
               minimum height=0.95cm, text width=2.7cm, inner sep=3pt, line width=0.7pt},
  badge/.style={circle, fill=#1, text=white, font=\bfseries\tiny, inner sep=1.6pt},
  pill/.style={rounded corners=4.5pt, fill=#1, text=white, font=\bfseries\tiny,
               inner xsep=5pt, inner ysep=2.2pt, align=center},
  flow/.style={->, >=stealth, line width=0.9pt, draw=black!75},
  cflow/.style={->, >=stealth, line width=0.8pt}
]
\definecolor{pahblue}{RGB}{47,82,177}
\definecolor{pahgreen}{RGB}{26,122,62}
\definecolor{pahamber}{RGB}{206,124,14}
\definecolor{pahred}{RGB}{176,32,32}
\definecolor{pahgray}{RGB}{120,120,120}

\draw[panel=pahblue] (-8.5,2.2) rectangle (8.5,-1.75);
\node[anchor=west, font=\footnotesize\bfseries, text=pahblue] at (-8.2,1.85)
  {Outer Loop: Closed Harness Optimization by the Optimization Agent};
\node[pill=pahblue, anchor=east] at (8.2,1.85) {development set only};

\node[stage=tongyi-purple, text width=1.55cm, fill=tongyi-purple!10] (best) at (-7.25,0.35)
  {\textbf{Current best}\\\textbf{harness}};

\node[stage=pahblue] (s1) at (-4.85,0.35) {\textbf{Attribution}\\mechanism-level failure analysis};
\node[stage=pahblue] (s2) at (-2.45,0.35) {\textbf{Proposal}\\one falsifiable\\change};
\node[stage=pahblue] (s3) at (-0.05,0.35) {\textbf{Implementation}\\code and prompt revision};
\node[stage=pahblue] (s4) at (2.35,0.35) {\textbf{Canary Gate}\\small\\targeted check};
\node[stage=pahblue] (s5) at (4.75,0.35) {\textbf{Full Evaluation}\\complete dev set run};

\node[badge=pahblue] at ([xshift=2pt,yshift=-1pt]s1.north west) {1};
\node[badge=pahblue] at ([xshift=2pt,yshift=-1pt]s2.north west) {2};
\node[badge=pahblue] at ([xshift=2pt,yshift=-1pt]s3.north west) {3};
\node[badge=pahblue] at ([xshift=2pt,yshift=-1pt]s4.north west) {4};
\node[badge=pahblue] at ([xshift=2pt,yshift=-1pt]s5.north west) {5};

\node[pill=pahgreen, text width=1.55cm] (acc) at (7.25,0.85) {Accept:\\new best harness};
\node[pill=pahred, text width=1.55cm] (rol) at (7.25,-0.15) {Rollback:\\keep the best};

\draw[flow] (best) -- (s1);
\draw[flow] (s1) -- (s2);
\draw[flow] (s2) -- (s3);
\draw[flow] (s3) -- (s4);
\draw[flow] (s4) -- (s5);
\draw[flow] (s5.east) -- (acc.west);
\draw[flow] (s5.east) -- (rol.west);
\draw[cflow, draw=black!60, dashed] (7.25,-0.42) -- (7.25,-1.25) -- (-7.25,-1.25) -- (best.south);
\node[font=\tiny\itshape, text=black!70, fill=pahblue!5, inner sep=1pt] at (0,-1.25)
  {next round starts from the current best harness; the dev metric alone decides acceptance};

\draw[flow, line width=1.1pt] (0,-1.75) -- (0,-2.35);
\node[anchor=west, font=\tiny\itshape, text=black!70] at (0.2,-2.05)
  {freeze code, prompts, and configuration; one-way evaluation on a disjoint test set};

\draw[panel=pahgreen] (-8.5,-2.35) rectangle (8.5,-7.35);
\node[anchor=west, font=\footnotesize\bfseries, text=pahgreen] at (-8.2,-2.7)
  {Frozen Optimized Harness at Runtime};
\node[pill=pahgreen, anchor=east] at (8.2,-2.7) {no Optimization Agent at runtime};

\node[stage=pahblue, text width=4.6cm, fill=pahblue!8] (env) at (0.4,-3.50)
  {\textbf{Web environment}\\search, visual retrieval, page reading};
\node[stage=pahblue, text width=2.5cm] (model) at (0.4,-5.00)
  {\textbf{Pistis-Agentic}};
\node[stage=pahgray, text width=1.6cm, fill=gray!8] (input) at (-6.95,-5.00)
  {Question\\and image};
\node[stage=tongyi-purple, text width=1.6cm, fill=tongyi-purple!10] (ans) at (7.0,-5.00)
  {\textbf{Final answer}};

\draw[flow, <->] (model) -- (env)
  node[midway, right=1pt, font=\tiny, text=black!70, align=left]
  {actions and observations,\\fixed interaction budget};
\draw[flow] (input) -- (model);
\draw[flow] (model) -- (ans)
  node[midway, above=0pt, font=\tiny, text=black!70] {budget-aware convergence};

\node[card=pahamber] (led) at (-4.4,-6.55)
  {\textbf{Candidate Ledger}\\candidates, evidence IDs, unmet constraints};
\node[card=pahgreen] (skl) at (0.4,-6.55)
  {\textbf{Search Skills}\\loaded only when the search is blocked};
\node[card=pahred] (chk) at (5.2,-6.55)
  {\textbf{Checkpoints and budget control}\\bounded prompts, reserved answer budget};

\draw[cflow, draw=pahamber] (led.north) -- ([xshift=-0.7cm]model.south);
\draw[cflow, draw=pahgreen] (skl.north) -- (model.south);
\draw[cflow, draw=pahred] (chk.north) -- ([xshift=0.7cm]model.south);
\end{tikzpicture}
\end{adjustbox}
\caption{\pahfix{Overview of Pistis-Auto-Harnessing (PAH). During development, an Optimization Agent runs a five-stage closed loop on a fixed development set: it attributes failures, proposes one falsifiable change, implements it, checks it with a small canary run, and evaluates it on the complete development set. A candidate replaces the current best harness only when the development metric improves; otherwise it is rolled back. The Optimized Harness is then frozen and evaluated once on a disjoint test set. At runtime, the Optimized Harness supports the frozen Pistis-Agentic model with a Candidate Ledger, conditionally loaded Search Skills, and bounded checkpoints with budget control, while the model itself makes every decision and writes the final answer.}}
\label{fig:auto_harness}
\end{figure*}
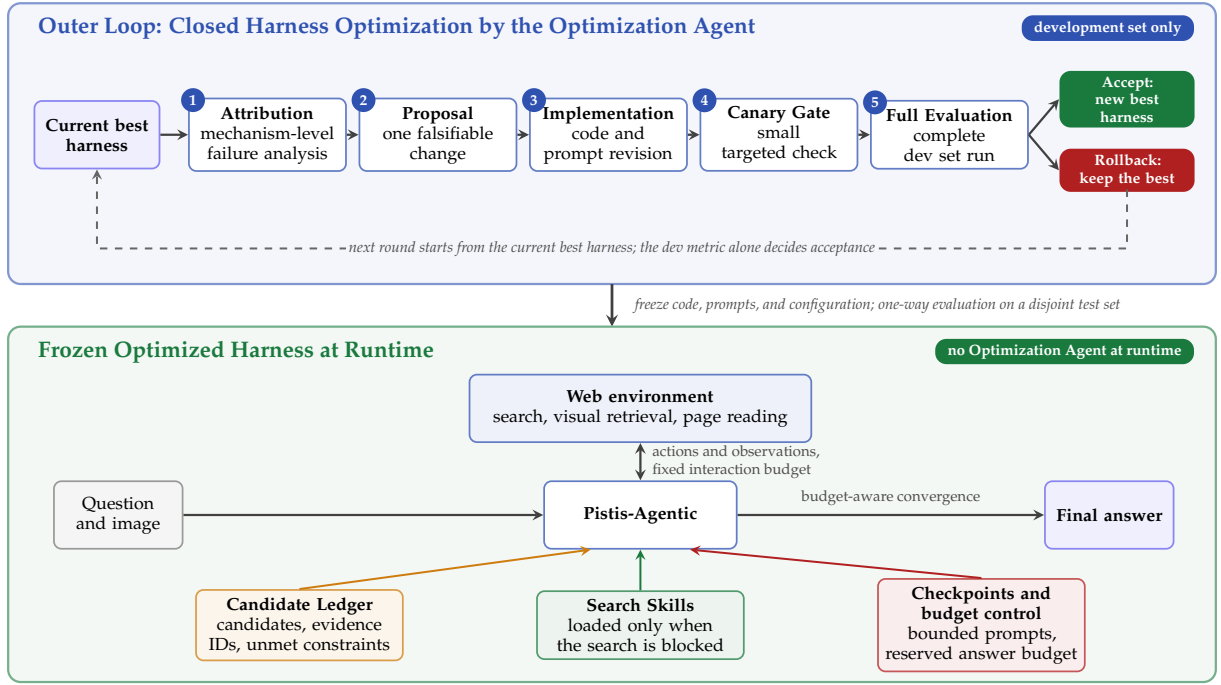

Auto-Harnessing is a system-level complement to IDRL for multimodal search. We instantiate it for Pistis-Agentic as \textbf{Pistis-Auto-Harnessing (PAH)}. Whereas IDRL improves the Pistis policy through parameter updates, PAH keeps the model, tool protocol, and evaluation entry point fixed and treats the surrounding inference harness as the optimization target. \pahfix{PAH denotes the optimization procedure rather than a particular runtime system. We call the initial system entering this procedure the Baseline Harness and the concrete system selected and frozen at the end of this application the Optimized Harness. Figure~\ref{fig:auto_harness} separates the development-time optimization procedure from the runtime behavior of its resulting harness.}

\subsubsection{\texorpdfstring{\pahfix{Auto-Harnessing Procedure}}{Auto-Harnessing Procedure}}

\pahfix{PAH optimizes the harness's code constraints, prompts, state representation, capability modules, routing rules, and workflow. Before the loop begins, we register the editable interface, the development metric, and a fixed environment-interaction budget. The Pistis-Agentic model remains frozen throughout, and no additional model is introduced into a task trajectory for routing, judging, voting, or answer selection. The Optimization Agent operates only during development and is absent from runtime once the Optimized Harness is ready.}

\pahfix{One harness revision follows a five-stage outer-loop cycle: attribution, proposal, implementation, canary gate, and full evaluation. The Optimization Agent reads the complete development-set run of the current best harness, attributes recurring failures at the mechanism level, and writes a falsifiable proposal specifying its trigger condition, expected state change, and rollback condition. Each round introduces one independently switchable change so that gains remain attributable. After implementation, a small canary matched to the target failure type must reach a preset target before the candidate enters a complete development-set evaluation under a fixed configuration. A candidate replaces the current best harness only when the pre-registered development metric improves; otherwise it is automatically rolled back, while mechanism activation rates and other process signals remain diagnostics only. The development and final test sets contain disjoint samples; the test set is used once after harness optimization and never feeds back into proposal generation or version selection.}

\subsubsection{\texorpdfstring{\pahfix{Baseline and Resulting Optimized Harness}}{Baseline and Resulting Optimized Harness}}

\pahfix{The components described below characterize the particular Optimized Harness obtained in this study, rather than constituting a fixed definition of PAH. Applying the same optimization procedure under a different task distribution, tool protocol, or budget may produce a different harness structure.}

\paragraph{\pahfix{Baseline Harness.}}
\pahfix{The Baseline Harness is a model-driven recurrent state machine. Given the question, image, prior messages, and previously returned images, the model either emits a \texttt{<tool\_call>} or a final \texttt{<answer>}; absent a valid tool call, the trajectory also terminates. At most one parsed tool is executed per turn, including web and image search, reverse image search, page summarization, and Python. Returned text is appended as a tool message and returned images are added to the multimodal context before the next model call. Malformed calls, repeated actions, and highly similar queries trigger retry or warning rules, while tool errors are returned as observations. A round or token-budget limit disables further tool calls and forces an answer from the accumulated context. This flexible loop leaves search decomposition and evidence retention to free-form next-action generation.}

\paragraph{\pahfix{Optimized Harness.}}
\pahfix{The frozen Optimized Harness produced in this study augments the baseline loop with a structured Candidate Ledger, conditionally loaded Search Skills, evidence-driven checkpoints, and budget-aware convergence. These mechanisms were proposed, tested, revised, or rejected by the PAH outer loop; they are properties of the resulting harness rather than manually prescribed components of the optimization method.}

\paragraph{\pahfix{Candidate Ledger.}}
\pahfix{The Candidate Ledger is a bounded, model-visible collection of structured candidate records injected into the model context. Each record contains a candidate entity or answer hypothesis, supporting and refuting evidence IDs, source relations, and unmet constraints. Every external result receives a stable evidence identifier, and a record can only cite valid identifiers, which keeps provenance traceable. The harness orders candidates with a deterministic scorer and surfaces evidence gaps, while the same reasoning model still makes every selection and writes the final answer; the ledger assists the model and never answers in its place. Evidence-driven checkpoints ask the model to record its current best candidate once enough external results exist, and later updates happen only when new evidence substantially changes, refutes, or adds a candidate. Checkpoints are bounded in number and budget, so they cannot form reflection loops. This design turns the choice between continuing to search and answering into a decision constrained jointly by evidence sufficiency, candidate gaps, and the remaining budget.}

\paragraph{\pahfix{Search Skills and adaptive workflow.}}
\pahfix{The resulting Search Skills are reusable operating procedures for recurring situations such as grounding visual candidates, disambiguating close candidates, tracing a relation chain, verifying the requested terminal field, repairing a failed query, or reconciling conflicting sources. Each skill records a trigger condition, action steps, success criteria, and a stop condition. It encodes a reusable procedure distilled from development trajectories, never an answer or a sample-specific rule.}

\pahfix{At runtime the skills are loaded conditionally. When the next action and its success condition are already clear, the model acts directly; a skill is loaded only when it would change the next evidence action, and each trajectory loads at most a small number of skills. An adaptive workflow coordinates three controls: advisory checkpoints after early evidence actions, a hard recovery step that loads the query-repair skill after a deterministic search failure, and the ledger checkpoint described above. The workflow also separates exploration from final answering. Near the interaction limit the harness stops opening new search branches, allows only a verification that can finish within the budget, then freezes further environment interaction and forces a final answer. All checkpoints have count and budget caps, so the model can proceed when it ignores a prompt or when the remaining budget is low.}

\pahfix{The whole runtime follows a minimal intervention principle: by default the harness preserves the original reasoning path, and it adds a candidate prompt, a routing step, or a recovery action only when a replayable structural state triggers it. The harness never constructs queries, candidates, or answers on behalf of the model, which keeps every state transition auditable from the trajectory.}

\pahadd{Runtime audits, transfer results, a Candidate Ledger case study, and limitations of the frozen Optimized Harness are provided in Appendix~\ref{app:auto_harness}.}

\subsubsection{\pahadd{Relation to Automated Agent Optimization}}

\pahadd{PAH shares with ADAS the use of an LLM-based optimizer to improve agent-system implementations rather than model parameters~\citep{hu2024adas}. Whereas ADAS emphasizes open-ended invention of code-defined agent architectures, PAH fixes the policy, task, environment interface, resource accounting, and evaluation protocol, then searches for auditable and reversible changes to the harness surrounding a single policy.}

\pahadd{Like AFlow, PAH treats code-level control flow rather than only a single instruction as an optimization target~\citep{zhang2024aflow}. AFlow searches graphs of LLM-calling nodes with Monte Carlo Tree Search; PAH requires neither a predefined multi-node graph nor a specific search algorithm, and may revise state representation, tool routing, bounded checkpoints, recovery, budget control, termination logic, and their associated prompts.}

\pahadd{PAH also resembles GEPA in using complete trajectories and natural-language reflection to diagnose failures and test revisions~\citep{agrawal2025gepa}. GEPA evolves prompts through reflective mutation and a Pareto frontier, whereas PAH may modify both deterministic code and prompts, evolves versions sequentially from the current best harness, and accepts a revision only when the fixed development metric improves. PAH therefore focuses on state-conditioned minimal intervention and auditable experimental governance for a frozen-policy runtime, rather than proposing a general open-ended, tree-search, or Pareto-evolution algorithm.}

\subsection{Infrastructure for Training, Evaluation and Inference}

The scalability of our framework rests on infrastructure for large-scale training, checkpoint-level evaluation, and efficient deployment, complementing the IDRL algorithm. We describe the three components in turn.

\subsubsection{Training}
We build the training infrastructure around reproducible, sandboxed environments with one-click replication via \texttt{uv}, reducing setup and migration costs across the SFT and RL stages. For continuous SFT, we adopt VeOmni~\citep{ma2025veomni} with a padding-free dynamic token-budget batching scheme, meta-device initialization, and rank-0-only checkpoint loading; combined with standard FSDP2 sharding, gradient checkpointing, mixed-precision training, FlashAttention-2, and optional Liger-Kernel, this gives over 80\% end-to-end speedup over a standard Hugging Face pipeline.

For on-policy distillation and RL, we adopt VeRL~\citep{sheng2024hybridflow}, whose asynchronous rollout pipeline decouples rollout generation from model optimization and executes them on separate resources, enabling sample generation and parameter updates to proceed in parallel. This removes the long-tail bottleneck of synchronous training, where updates are stalled by the slowest rollout. Using vLLM as the rollout engine, whose high-throughput serving and KV-cache reuse amortize decoding overhead across batched rollouts, this design yields an overall $\sim$2$\times$ end-to-end speedup over conventional synchronous RL pipelines.

\begin{figure}[t]
    \centering
    \includegraphics[width=0.8\textwidth]{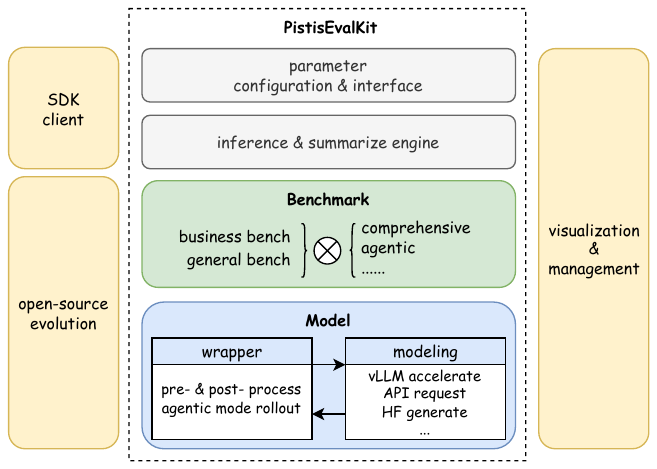}
    \caption{Architecture of PistisEvalKit.}
    \label{fig:pistis_evalkit_arch}
\end{figure}

\subsubsection{Evaluation}
We evaluate Pistis with \textbf{PistisEvalKit}, built on the open-source VLMEvalKit~\citep{duan2024vlmevalkit}. As shown in Figure~\ref{fig:pistis_evalkit_arch}, it adopts a modular design with two decoupled components: a \textbf{Model} module that separates standardized pre/post-processing (the Wrapper, including agentic workflows such as tool calling) from inference backends (the Modeling layer, covering vLLM~\citep{vllm}, Hugging Face Transformers~\citep{transformers}, and API services), and a \textbf{Benchmark} module that remains compatible with VLMEvalKit benchmarks while extending to internal suites (e.g., Pistis Benchmark) and agentic tasks.

For all publicly available models evaluated in our environment, we follow the inference configurations officially recommended by their respective model developers, including the reasoning mode and decoding parameters. Qwen3.8-27B is evaluated with the officially recommended \texttt{xhigh} reasoning effort, while the other reasoning-enabled baselines use their recommended \texttt{thinking} configurations. For Pistis models, we use temperature $0.0$ and top-$p$ $1.0$ to minimize sampling variance. Results for Step3-VL-10B are taken directly from its technical report rather than reproduced in our evaluation environment.

Two capabilities are tailored to our development workflow. First, an \textbf{automated training-evaluation loop}: an SDK-based client orchestrates large-scale evaluation and automatically evaluates checkpoints during training, enabling fine-grained monitoring and faster iteration. Second, \textbf{native agentic benchmarking}: PistisEvalKit supports multi-turn reasoning and tool calling via DeepEyesV2~\citep{hong2025deepeyesv2} with an extensible tool ecosystem. Configuration uses a Hydra-based YAML system with Pydantic~\citep{pydantic} typing for one-command execution of large benchmark suites.

\subsubsection{Inference}
\rev{We optimize Pistis online inference on top of vLLM at the system, scheduling, and kernel levels, improving inference speed by over 100\% in some prefill scenarios. Two optimizations follow vLLM's design and are tuned for \revdel{Qwen3-VL} Qwen3.5/Qwen3.6 visual-language workloads: an \textbf{asynchronous pipeline} that overlaps CPU pre/post-processing with GPU compute by converting synchronous operators, including H2D copies and boolean-mask operations, to asynchronous forms, reaching up to 99\% GPU utilization with multi-stream; and a \textbf{multi-process} frontend/backend split that relieves the Python GIL for CPU-bound request handling, tokenization, and image processing, using shared memory for large multimodal payloads.}

\rev{At the kernel level, we further apply \textbf{parameter alignment}. NVIDIA's Tensor Memory Accelerator (TMA) on the Hopper architecture requires parameter dimensions to be multiples of 128 to schedule its high-performance operators. We identify unaligned ViT parameters in the \revdel{Qwen3-VL} Qwen3.5/Qwen3.6-based models and pad the affected dimensions to 128-aligned sizes, enabling high-performance operators on H20 GPUs.}

\begin{table*}[t]
\centering
\caption{Performance of Pistis-27B, Pistis-9B, and representative multimodal large language models, including Qwen3.6-27B, Qwen3.8-27B, Qwen3.5-9B, Step3-VL-10B, Qwen3-VL-8B~\citep{qwen3vl}, Keye-VL-1.5~\citep{keye2025keyevl15}, and InternVL3.5~\citep{wang2025internvl35}, on visual benchmarks. Results for Step3-VL-10B are taken directly from its technical report, while all other results are obtained using our unified evaluation environment. The final two rows report averages separately over 24 non-grounding benchmarks and six grounding benchmarks.}
\label{tab:results_small}
\begin{adjustbox}{max width=1\textwidth}
\setlength{\tabcolsep}{1pt}
\begin{NiceTabular}{c|l|w{c}{1.8cm}w{c}{1.8cm}w{c}{1.8cm}|w{c}{1.8cm}w{c}{1.8cm}w{c}{1.8cm}w{c}{1.8cm}w{c}{1.8cm}w{c}{1.8cm}}
\toprule
\multirow{2}{*}{\textbf{Category}} & \multirow{2}{*}{\textbf{Benchmark}} &
\Block{1-1}{\textbf{Pistis}\\\textbf{27B}} &
\Block{1-1}{\textbf{Qwen3.6}\\\textbf{27B}} &
\Block{1-1}{\textbf{Qwen3.8}\\\textbf{27B}} &
\Block{1-1}{\textbf{Pistis}\\\textbf{9B}} &
\Block{1-1}{\textbf{Qwen3.5}\\\textbf{9B}} &
\Block{1-1}{\textbf{Step3-VL}\\\textbf{10B}} &
\Block{1-1}{\textbf{Qwen3-VL}\\\textbf{8B}} &
\Block{1-1}{\textbf{Keye-VL-1.5}\\\textbf{8B}} &
\Block{1-1}{\textbf{InternVL3.5}\\\textbf{8B}} \\
& & {\scriptsize thinking} & {\scriptsize thinking} & {\scriptsize xhigh} & {\scriptsize thinking} & {\scriptsize thinking} & {\scriptsize thinking} & {\scriptsize thinking} & {\scriptsize thinking} & {\scriptsize thinking} \\
\midrule
\Block{4-1}{STEM \\ Puzzle}
& MMMU$_{val}$ & 81.0 & 82.0 & \textbf{82.3} & 76.6 & 78.0 & \textbf{78.11} & 71.6 & 71.4 & 73.4 \\
& ScienceQA$_{val}$ & \textbf{99.5} & 98.5 & 98.7 & \textbf{99.2} & 98.1 & - & 95.8 & 97.6 & 96.7 \\
& MathVista$_{\text{mini}}$ & \textbf{87.7} & 87.5 & 87.0 & \textbf{86.8} & 85.3 & 83.97 & 79.2 & 81.2 & 80.8 \\
& MathVerse$_{\text{mini}}$ & \textbf{87.5} & 87.0 & 87.2 & \textbf{85.9} & 84.7 & 75.73 & 73.1 & 70.9 & 61.9 \\
\midrule
\Block{4-1}{General \\ VQA}
& RealWorldQA & 84.6 & 84.2 & \textbf{85.8} & \textbf{81.4} & 81.0 & 74.44 & 73.2 & 73.2 & 68.6 \\
& MMStar & 80.5 & \textbf{81.3} & 80.1 & 79.7 & 78.9 & 77.48 & 75.3 & \textbf{80.3} & 66.0 \\
& MM-Vet & \textbf{77.9} & 76.9 & 77.1 & \textbf{78.1} & 75.2 & - & 71.1 & 73.2 & 70.0 \\
& MME & \textbf{91.6} & 89.1 & 87.3 & \textbf{89.8} & 89.3 & - & 84.6 & 86.0 & 84.6 \\
\midrule
\Block{2-1}{Alignment}
& HallusionBench & 69.4 & 67.7 & \textbf{69.6} & \textbf{69.3} & 66.8 & 64.91 & 61.8 & 64.2 & 59.1 \\
& MMVP & 81.7 & \textbf{85.0} & 84.0 & 79.7 & \textbf{83.0} & 68.16 & 78.7 & 78.7 & 72.7 \\
\midrule
\Block{8-1}{Document \\ Understanding}
& TextVQA$_{val}$ & 89.0 & \textbf{89.3} & \textbf{89.3} & 88.6 & \textbf{89.0} & - & 85.9 & 86.2 & 83.2 \\
& AI2D$_{\text{test}}$ & \textbf{93.2} & 92.6 & 92.6 & \textbf{92.4} & 91.2 & 89.35 & 85.1 & 90.0 & 83.6 \\
& ChartQA$_{\text{test}}$ & 84.8 & 86.1 & \textbf{86.4} & 84.4 & \textbf{86.0} & - & 84.2 & 82.7 & 78.1 \\
& InfoVQA$_{val}$ & 93.6 & 93.8 & \textbf{93.9} & 91.3 & \textbf{91.4} & - & 84.9 & 77.6 & 79.1 \\
& DocVQA$_{val}$ & \textbf{96.0} & 95.7 & 95.9 & \textbf{95.8} & 94.6 & - & 93.2 & 92.5 & 92.3 \\
& OCRBench & \textbf{91.2} & 88.0 & 85.3 & \textbf{91.1} & 88.9 & 86.75 & 83.1 & 86.6 & 84.0 \\
& CharXiv(DQ)$_{val}$ & 95.0 & \textbf{95.3} & 94.8 & \textbf{93.5} & 93.3 & - & 89.3 & 78.8 & 80.5 \\
& CharXiv(RQ)$_{val}$ & \textbf{80.7} & 77.7 & \textbf{80.7} & 72.5 & \textbf{74.4} & 59.52 & 54.3 & 46.2 & 49.3 \\
\midrule
\Block{4-1}{Spatial \\ Grounding}
& RefCOCO$_{testA}$ & \textbf{95.8} & 94.4 & 94.6 & \textbf{95.5} & 92.8 & - & 93.3 & 85.3 & 94.7 \\
& RefCOCO$_{testB}$ & \textbf{91.6} & 89.8 & 89.5 & \textbf{91.1} & 86.8 & - & 87.4 & 74.5 & 88.7 \\
& RefCOCO+$_{testA}$ & \textbf{94.0} & 92.3 & 92.5 & \textbf{93.9} & 89.7 & - & 90.2 & 82.3 & 92.4 \\
& RefCOCO+$_{testB}$ & \textbf{86.7} & 84.7 & 85.1 & \textbf{86.7} & 80.5 & - & 80.7 & 68.7 & 82.4 \\
\midrule
\Block{2-1}{Temporal \\ Grounding}
& Charades-STA$_{64frame}$ & \textbf{63.2} & 56.9 & 34.5 & \textbf{63.6} & 54.1 & - & 57.5 & 22.2 & 23.8 \\
& TACoS$_{128frame}$ & \textbf{51.5} & 40.3 & 35.0 & \textbf{49.8} & 34.1 & - & 34.4 & 3.0 & 4.4 \\
\midrule
\Block{1-1}{Multi-Image}
& BLINK & 73.6 & 74.0 & \textbf{80.7} & \textbf{71.6} & 70.0 & 66.79 & 63.3 & 56.1 & 57.7 \\
\midrule
\Block{4-1}{Video \\ Understanding}
& MVBench$_{8frame}$ & 70.6 & 70.4 & \textbf{71.9} & \textbf{68.9} & 67.8 & - & 66.2 & 56.9 & 67.5 \\
& TempCompass$_{8frame}$ & 83.1 & 81.8 & \textbf{85.8} & \textbf{81.0} & 79.8 & - & 76.6 & 72.8 & 72.1 \\
& MLVU$_{64frame}$ & 73.9 & \textbf{74.9} & 71.8 & 72.7 & 71.9 & - & 68.0 & \textbf{75.0} & 71.0 \\
& Video-MME$_{64frame}$ & 74.0 & 75.2 & \textbf{76.1} & 70.1 & 68.2 & - & 65.2 & \textbf{73.0} & 65.4 \\
\midrule
\Block{1-1}{Multilingual}
& MTVQA$_{test}$ & \textbf{34.6} & \textbf{34.6} & 33.3 & 33.8 & 32.1 & - & 26.7 & 25.0 & \textbf{35.2} \\
\midrule
\Block{2-1}{\textbf{Summary}}
& \textbf{Non-grounding Avg. (24)} & 82.3 & 82.0 & \textbf{82.4} & \textbf{80.6} & 80.0 & - & 74.6 & 74.0 & 72.2 \\
& \textbf{Grounding Avg. (6)} & \textbf{80.5} & 76.4 & 71.9 & \textbf{80.1} & 73.0 & - & 73.9 & 56.0 & 64.4 \\
\bottomrule
\end{NiceTabular}
\end{adjustbox}
\end{table*}
\section{Experiments}
\label{sec:exp}

\subsection{Comparison with Public Benchmarks}
\subsubsection{Perception and Reasoning Tasks}

As shown in Table~\ref{tab:results_small}, we report aggregate results separately over 24 non-grounding benchmarks and six spatial or temporal grounding benchmarks. On the non-grounding subset, Pistis-27B obtains an average score of 82.3, slightly outperforming Qwen3.6-27B (82.0) while remaining essentially on par with Qwen3.8-27B (82.4). Across all 30 benchmarks, Pistis-27B records 19 wins, 10 losses, and one tie against Qwen3.6-27B, and 18 wins, 11 losses, and one tie against Qwen3.8-27B. The improvements are therefore broad but not uniform. Pistis-27B remains weaker on several benchmarks, including MMMU, MMVP, and ChartQA, and trails Qwen3.8-27B by 7.1 points on BLINK. Its largest margins are concentrated in spatial and temporal grounding, where its six-benchmark average reaches 80.5, compared with 76.4 for Qwen3.6-27B and 71.9 for Qwen3.8-27B.

At the 9B scale, Pistis-9B achieves a non-grounding average of 80.6, exceeding Qwen3.5-9B by 0.6 points, and records 24 wins and six losses over the full set of 30 benchmarks. It improves on Qwen3.5-9B across STEM reasoning, document understanding, and most video-understanding tasks, while remaining weaker on MMVP (79.7 vs.\ 83.0). As with the 27B model, the largest gains occur on grounding benchmarks: Pistis-9B averages 80.1 over the six grounding tasks, compared with 73.0 for Qwen3.5-9B. These results indicate that Pistis provides consistent but generally moderate gains outside grounding, together with substantially stronger spatial and temporal grounding capability.

\subsubsection{Agentic Tasks}
\begin{table*}[t]
\centering
\caption{Agentic benchmark performance comparison among Pistis-27B Agentic, Pistis-9B Agentic, Qwen3.6-27B, Qwen3.8-27B, Qwen3.5-9B, DeepEyesV2-7B~\citep{hong2025deepeyesv2}, and Thyme-7B~\citep{zhang2025thyme}. All results in this table are evaluated with prompts that expose the tool schema. \pahfix{Consequently,} scores on benchmarks overlapping with Table~\ref{tab:results_small} are not directly comparable because the prompting configurations differ. MMSearch combines the multimodal and text-only subsets; BrowseComp-VL similarly combines Level 1 and Level 2. We use a general-purpose ReAct-based harness in which the agent dynamically selects and sequences atomic tool calls. This differs from the official MMSearch's fixed search workflow and VDR-testmini's composite tools, which bundle operations such as image search and cropping. We report the main score on PinchBench averaged over five runs. }
\label{tab:results_agentic}
\begin{adjustbox}{max width=1\textwidth}
\begin{NiceTabular}{c|l|ccc|cccc}
\toprule
Category & Benchmark & \shortstack{Pistis 27B \\ Agentic} & \shortstack{Qwen3.6 \\ 27B} & \shortstack{Qwen3.8 \\ 27B xhigh} & \shortstack{Pistis 9B \\ Agentic} & \shortstack{Qwen3.5 \\ 9B} & \shortstack{DeepEyesV2 \\ 7B} & \shortstack{Thyme \\ 7B} \\
\midrule

\multirow{3}{*}{\shortstack{Chart \\ Understanding}}
& ChartQA$_{\text{test}}$ & 85.0 & \textbf{86.2} & 78.2 & 83.5 & 83.7 & \textbf{88.4} & 86.1 \\
& CharXiv(DQ)        & \textbf{95.0} & \textbf{95.0} & \textbf{95.0} & \textbf{92.3} & \textbf{92.3} & 78.6 & - \\
& CharXiv(RQ)        & 77.7 & 75.5 & \textbf{80.4} & \textbf{70.2} & 64.6 & 48.9 & - \\
\midrule

\multirow{7}{*}{\shortstack{Real-World \\ Perception}}
& V*                  & 94.2 & \textbf{94.8} & 91.1 & \textbf{93.7} & 91.1 & 81.8 & 82.2 \\
& TreeBench           & 59.8 & 51.1 & \textbf{65.7} & \textbf{55.3} & 52.3 & 42.5 & - \\
& OCRBench            & \textbf{87.1} & 86.8 & 81.5 & 87.0 & \textbf{88.1} & - & - \\
& SeedBench-2 Plus    & \textbf{76.9} & 76.3 & 76.6 & 75.1 & 74.4 & \textbf{78.6} & - \\
& HRBench4K           & \textbf{91.3} & 91.0 & 88.5 & \textbf{89.0} & 87.9 & 77.9 & 77.0 \\
& HRBench8K           & \textbf{89.6} & 87.6 & 89.5 & 84.9 & \textbf{85.1} & 73.8 & 72.0 \\
& MME-RealWorld-Lite  & 63.3 & 61.5 & \textbf{66.1} & 63.6 & 59.2 & \textbf{64.9} & 64.8 \\
\midrule

\multirow{3}{*}{\shortstack{Multimodal \\ Reasoning}}
& MathVista$_{\text{mini}}$ & \textbf{87.7} & 86.1 & 87.3 & \textbf{83.5} & 82.0 & 71.9 & 70.0 \\
& MathVerse$_{\text{mini}}$ & 86.9 & 85.8 & \textbf{88.1} & \textbf{84.1} & 79.8 & 52.7 & - \\
& LogicVista                 & 81.0 & 78.3 & \textbf{83.7} & \textbf{70.7} & 69.4 & 48.7 & 49.0 \\
\midrule

\multirow{4}{*}{\shortstack{Search-\\Oriented}}
& BrowseComp-VL & \textbf{57.2} & 48.4 & 54.6 & \textbf{53.2} & 44.8 & - & - \\
& MMSearch      & \textbf{78.0} & 74.7 & 74.7 & \textbf{72.7} & 64.7 & 63.7 & - \\
& VDR-testmini  & \textbf{26.8} & 23.6 & 26.6 & \textbf{24.8} & 21.6 & - & - \\
& LiveVQA       & 84.7 & 75.0 & \textbf{88.3} & \textbf{82.7} & 71.3 & - & - \\
\midrule
Claw-Style
& PinchBench    & 86.5 & 85.7 & \textbf{87.8} & \textbf{77.6} & 74.8 & - & - \\
\midrule
Overall & Average & \textbf{78.3} & {75.7} & 78.0 & \textbf{74.7} & {71.5} & - & - \\
\bottomrule

\end{NiceTabular}
\end{adjustbox}
\end{table*}

As shown in Table~\ref{tab:results_agentic}, Pistis-27B-Agentic and Pistis-9B-Agentic obtain higher overall averages than their corresponding Qwen base models across the 18 reported agentic benchmarks. Pistis-27B-Agentic reaches 78.3, compared with 75.7 for Qwen3.6-27B, while Pistis-9B-Agentic reaches 74.7, compared with 71.5 for Qwen3.5-9B. Multimodal search is a major source of improvement over these base models. Pistis-27B-Agentic exceeds Qwen3.6-27B by 8.8 points on BrowseComp-VL, 3.3 points on MMSearch, 3.2 points on VDR-testmini, and 9.7 points on LiveVQA. The corresponding gains for Pistis-9B-Agentic over Qwen3.5-9B are 8.4, 8.0, 3.2, and 11.4 points, respectively. On PinchBench, the reported mean scores increase from 85.7 to 86.5 at the 27B scale and from 74.8 to 77.6 at the 9B scale. Beyond these interaction benchmarks, Pistis-27B-Agentic also exceeds Qwen3.6-27B by 8.7 points on TreeBench and 2.0 points on HRBench8K.

Pistis-27B-Agentic and Qwen3.8-27B obtain similar overall scores (78.3 vs.\ 78.0), with a small numerical advantage for Pistis-27B-Agentic and task-dependent differences. Pistis-27B-Agentic scores higher on ChartQA (85.0 vs.\ 78.2), V* (94.2 vs.\ 91.1), OCRBench (87.1 vs.\ 81.5), HRBench4K (91.3 vs.\ 88.5), BrowseComp-VL (57.2 vs.\ 54.6), and MMSearch (78.0 vs.\ 74.7), while Qwen3.8-27B scores higher on tasks including TreeBench (65.7 vs.\ 59.8), LiveVQA (88.3 vs.\ 84.7), and PinchBench (87.8 vs.\ 86.5). Taken together, these results demonstrate the effectiveness of our post-training recipe: starting from Qwen3.6-27B, Pistis-27B-Agentic raises the overall benchmark average from 75.7 to 78.3, achieving competitive aggregate performance against Qwen3.8-27B while retaining different strengths across individual tasks.

\subsection{\texorpdfstring{\pahfix{Held-Out Evaluation of the PAH-Optimized Harness}}{Held-Out Evaluation of the PAH-Optimized Harness}}
\label{sec:auto_harness_exp}

\pahfix{We evaluate the frozen \textbf{Optimized Harness} produced by Pistis-Auto-Harnessing (PAH), using Pistis-27B-Agentic as the frozen policy. During development, the outer loop described in Section~\ref{sec:auto_harness} runs on 100 development instances that share the source and distribution of the test set but contain no overlapping samples. All attribution, proposals, canary runs, and version selection use only this development set. The selected harness is then frozen, including its code, prompts, and configuration, and evaluated once on the VDR-testmini. In the primary matched-budget comparison, the Optimized Harness and the Baseline Harness share the same limit of 15 environment interactions per trajectory, counting search, visual retrieval, and page reading; the 10- and 30-interaction Baseline runs are budget references only.}

\pahfix{As shown in Table~\ref{tab:auto_harness_results}, the frozen Optimized Harness reaches 28.6\% accuracy, compared with 26.8\% for the Baseline Harness, a gain of 1.8 percentage points. The average number of actual environment interactions is nearly unchanged (8.60 versus 8.68), so the gain comes from how the fixed budget is spent. We attribute the improvement to the frozen harness as a whole, covering the Candidate Ledger, the Search Skills, the adaptive workflow, and the budget design; the test set took no part in any version selection. The result provides descriptive evidence for a complementary system-level contribution: IDRL supplies the trained agentic policy, whereas PAH organizes how that fixed policy retrieves, preserves, and adjudicates evidence.} \pahadd{The same frozen harness also transfers to Seed-2.1-turbo and GPT-5.5, yielding gains of 1.4 and 4.0 percentage points, respectively (Appendix~\ref{app:pah_policy_transfer}).}

\pahadd{A budget sweep provides additional descriptive context. With maximum interaction limits of 10, 15, and 30, the Baseline Harness reaches 25.2\%, 26.8\%, and 28.4\% accuracy while using 6.55, 8.68, and 11.66 average interactions, respectively. The Optimized Harness reaches 28.6\% under the matched 15-interaction limit while using 8.60 interactions. It is numerically 0.2 percentage points higher than the 30-interaction Baseline while using 26.2\% fewer interactions, suggesting that the observed difference is not attributable solely to a larger search budget.}

\begin{table*}[t]
\centering
\caption{\pahfix{Frozen Optimized Harness versus Baseline Harness on the VDR-testmini, with Pistis-27B-Agentic as the frozen policy. The primary comparison uses the same limit of 15 environment interactions per trajectory, and the test set took no part in version selection.} \pahadd{The 10- and 30-interaction Baselines are included only as budget references.}}
\label{tab:auto_harness_results}
\begin{adjustbox}{max width=\textwidth}
\begin{tabular}{lcccc}
\toprule
\textbf{Harness} & \pahadd{\textbf{Max. env. interactions}} & \textbf{Accuracy (\%)} & \textbf{Correct/Total} & \textbf{Avg. env. interactions} \\
\midrule
\pahadd{Baseline Harness} & \pahadd{10} & \pahadd{25.2} & \pahadd{126/500} & \pahadd{6.55} \\
Baseline Harness & \pahadd{15} & 26.8 & 134/500 & 8.68 \\
\pahadd{Baseline Harness} & \pahadd{30} & \pahadd{28.4} & \pahadd{142/500} & \pahadd{11.66} \\
\midrule
Optimized Harness & \pahadd{15} & \textbf{28.6} & \textbf{143/500} & 8.60 \\
\bottomrule
\end{tabular}
\end{adjustbox}
\end{table*}

\paragraph{\pahadd{Frozen cross-benchmark transfer.}}
\pahadd{We directly transfer the same frozen Optimized Harness to three additional search benchmarks without using their examples to optimize the harness or select a version. As shown in Table~\ref{tab:auto_harness_transfer}, performance improves on MMSearch, BrowseComp-VL, and LiveVQA by 0.4, 1.7, and 1.0 points, respectively. The consistently positive direction provides evidence that the resulting candidate-maintenance, evidence-organization, and convergence procedures transfer beyond VDR-testmini, although all evaluated tasks remain within multimodal search.}

\begin{center}
\begin{minipage}{\linewidth}
\centering
\captionof{table}{\pahadd{Zero-target-tuning transfer of the frozen Optimized Harness. The harness is not revised or selected on any of the three target benchmarks.}}
\label{tab:auto_harness_transfer}
\begin{adjustbox}{max width=\textwidth}
\begin{tabular}{lccc}
\toprule
\textbf{Benchmark} & \textbf{Baseline Harness} & \textbf{Optimized Harness} & \textbf{Gain} \\
\midrule
MMSearch      & 78.0 & \textbf{78.4} & +0.4 \\
BrowseComp-VL & 57.2 & \textbf{58.9} & +1.7 \\
LiveVQA       & 84.7 & \textbf{85.7} & +1.0 \\
\bottomrule
\end{tabular}
\end{adjustbox}
\end{minipage}
\end{center}
\pahadd{Together, these evaluations isolate harness-level gains for fixed policies; we next return to the model-level contribution and ablate the components of IDRL.}

\subsection{Ablation Study}

\begin{figure}[t]
     \centering
     \begin{subfigure}[b]{0.48\textwidth}
         \centering
         \includegraphics[width=\textwidth]{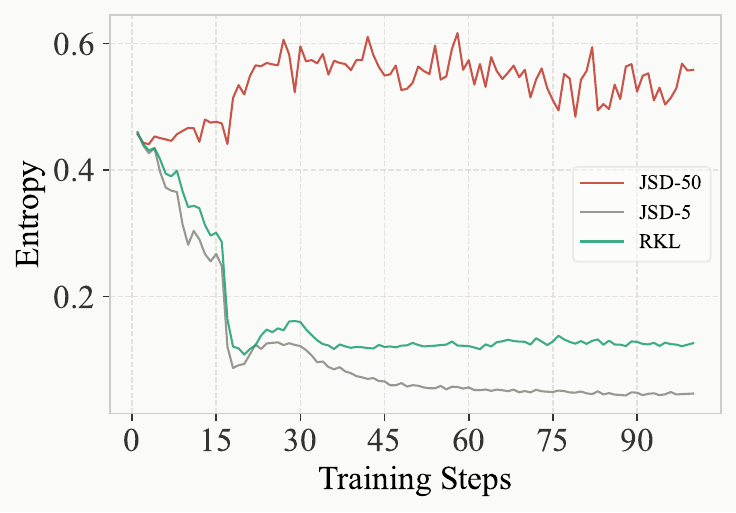}
         \caption{}
     \end{subfigure}
     \hfill
     \begin{subfigure}[b]{0.48\textwidth}
         \centering
         \includegraphics[width=\textwidth]{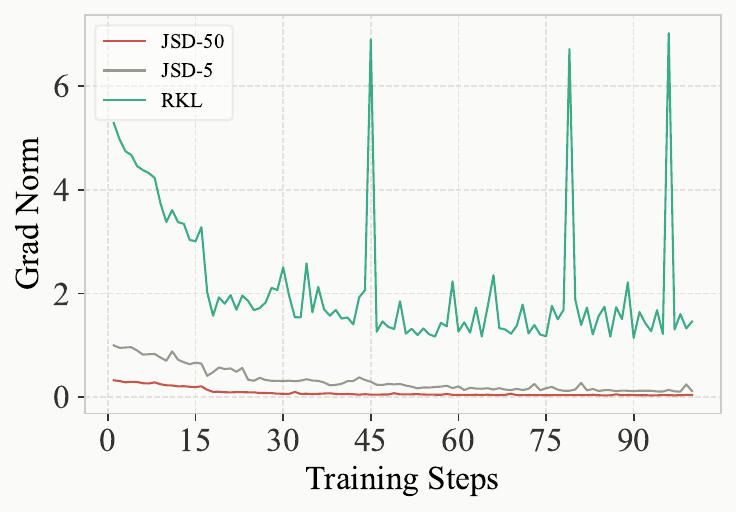}
         \caption{}
     \end{subfigure}
     
     \caption{Effect of OPD objective design on training stability. We compare entropy (a) and gradient norm (b) across RKL, JSD-5, and JSD-50. Broader top-$k$ JSD preserves higher policy entropy and yields smoother gradients, while RKL and narrow-support JSD exhibit entropy collapse or instability.}
     \label{fig:OPD_analysis}
\end{figure}

\noindent \textbf{Impact of OPD variants.}
Figure~\ref{fig:OPD_analysis} compares the training dynamics of entropy and gradient norm across three OPD variants: Reverse KL (RKL), JSD-5, and JSD-50. RKL, which computes the KL divergence solely from the current predicted token~\citep{lu2025onpolicydistillation}, exhibits severe training instability with large, irregular gradient-norm spikes. This mode-seeking behavior also concentrates probability mass on a narrow set of tokens, ultimately leading to entropy collapse as shown by the low entropy values after convergence. Replacing RKL with our JSD formulation substantially alleviates gradient instability: both JSD-5 and JSD-50 maintain much smoother gradient norms. However, JSD-5, which restricts the divergence computation to the top-5 tokens, still suffers from entropy collapse, suggesting that five candidate tokens are insufficient to prevent over-concentration.
In contrast, JSD-50 expands the divergence computation to the top-50 tokens, providing a broader supervisory signal that prevents over-concentration of probability mass. Crucially, this wider token coverage acts as a threshold: \textbf{once enough candidate tokens are covered, the model entropy rises and stabilizes at a higher level}. This elevated entropy is not merely a training artifact; it reflects more diverse output distributions that are essential for downstream reinforcement learning, where greater sampling diversity translates to richer exploration and more effective policy optimization. Based on this analysis, we adopt top-50 tokens for JSD computation in our framework.

\begin{figure}[t]
     \centering
     \begin{subfigure}[b]{0.48\textwidth}
         \centering
         \includegraphics[width=\textwidth]{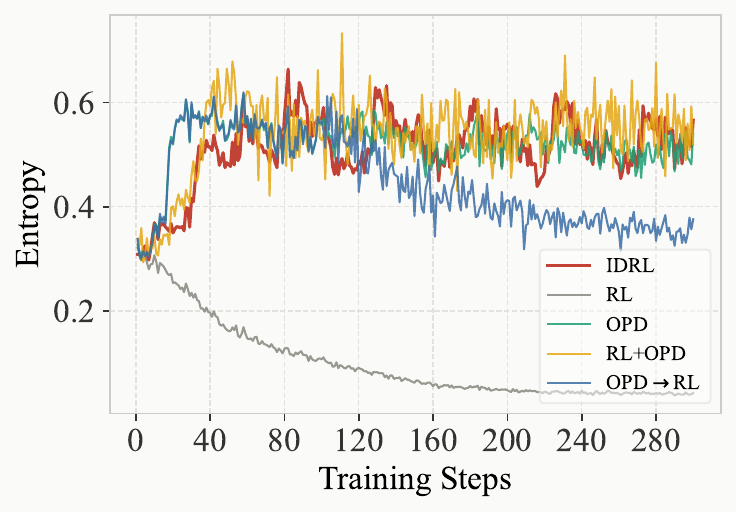}
         \caption{}
         \label{fig:idrl_entropy}
     \end{subfigure}
     \hfill
     \begin{subfigure}[b]{0.48\textwidth}
         \centering
         \includegraphics[width=\textwidth]{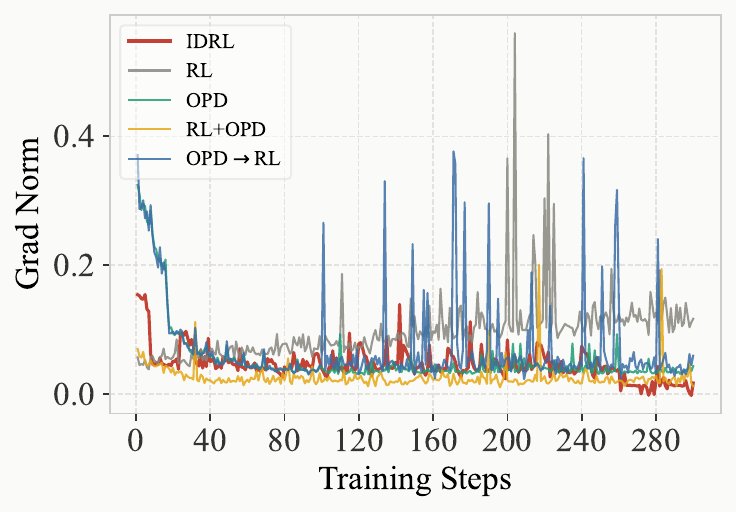}
         \caption{}
         \label{fig:idrl_grad_norm}
     \end{subfigure}
     
     \caption{Effect of interleaving OPD and RL during post-training. Compared with vanilla RL, vanilla OPD, and the joint RL+OPD objective, IDRL shows the expected phase-wise entropy pattern (a) and more stable gradient norms (b), indicating that alternating the two objectives reduces optimization interference.}
     \label{fig:IDRL_analysis}
\end{figure}

\noindent \textbf{Effect of IDRL.}
Starting from the same SFT checkpoint, we compare pure RL (SAPO), pure OPD, a two-stage sequential pipeline that applies OPD followed by RL (OPD$\rightarrow$RL), their joint optimization (RL+OPD with $\mathcal{L}_{\text{joint}}=-\mathcal{J}_{\text{RL}}+\alpha\mathcal{L}_{\text{OPD}}$), and our interleaved variant (IDRL).
The training dynamics in Figure~\ref{fig:IDRL_analysis} show that pure RL undergoes steady entropy collapse, whereas OPD, RL+OPD, and IDRL maintain substantially higher entropy. IDRL exhibits phase-wise entropy variation consistent with its alternating schedule: entropy tends to decrease during RL phases and recover during OPD phases. In terms of optimization stability, pure RL develops a rising gradient norm and severe late-stage spikes, while the methods incorporating OPD maintain lower and more bounded gradient norms. These results suggest that interleaving preserves the complementary effects of RL and OPD without forcing their potentially conflicting gradients into every update.
Table~\ref{tab:ablation_idrl} provides the corresponding downstream results over 18 benchmarks. IDRL achieves the highest reported overall average of 74.7, compared with 74.1 for sequential OPD$\rightarrow$RL and approximate averages of 74.0 for pure RL, 74.1 for joint RL+OPD, and 73.4 for pure OPD. The comparisons with the joint and sequential variants examine two distinct alternatives to interleaving. Relative to joint RL+OPD, IDRL obtains higher scores in all five categories, with differences of 0.7 on Chart Understanding, 0.7 on Real-World Perception, 0.5 on Multimodal Reasoning, 0.5 on Search-Oriented tasks, and 0.1 on PinchBench. This pattern favors alternating the objectives over combining them within every update in the evaluated setting. Relative to sequential OPD$\rightarrow$RL, IDRL obtains higher scores in four categories, with the largest numerical gain on Search-Oriented tasks (+1.9) and a gain of 0.4 on PinchBench, while scoring slightly lower on Real-World Perception (78.4 vs.\ 78.5). This comparison favors repeated alternation over a single transition from distillation to RL in terms of aggregate performance, though not in every category. Across all evaluated variants, IDRL ranks highest in Chart Understanding, Multimodal Reasoning, Search-Oriented tasks, and PinchBench. Together with the observed training dynamics, these results support interleaving as a promising way to combine RL and OPD, without establishing that reduced gradient interference or improved exploration alone explains the downstream differences.

\begin{table*}[t]
\centering
\caption{Comparison of IDRL against vanilla RL, OPD, the sequential OPD-then-RL pipeline (OPD$\rightarrow$RL), and joint RL+OPD optimization on Pistis-9B-Agentic across agentic benchmark categories. All models start from the SFT checkpoint. AVG gives equal weight to each of the 18 individual benchmarks.}
\label{tab:ablation_idrl}
\renewcommand{\arraystretch}{1.3}
\begin{tabular}{lccccc!{\vrule}c}
\toprule
\textbf{Category} & \textbf{SFT} & \textbf{RL} & \textbf{OPD} & \textbf{OPD$\rightarrow$RL} & \textbf{RL+OPD} & \textbf{IDRL} \\
\midrule
Chart Understanding    & 81.0 & 81.5 & 79.6 & 81.2 & 81.3 & \textbf{82.0} \\
Real-World Perception  & 77.5 & 77.8 & \textbf{78.5} & \textbf{78.5} & 77.7 & 78.4 \\
Multimodal Reasoning   & 78.5 & 79.0 & 77.8 & 78.9 & 78.9 & \textbf{79.4} \\
Search-Oriented        & 57.1 & 57.3 & 56.2 & 56.5 & 57.9 & \textbf{58.4} \\
Claw-Style             & 77.1 & 77.5 & 74.4 & 77.2 & 77.5 & \textbf{77.6} \\
\midrule
\textbf{AVG}           & 73.7 & 74.0 & 73.4 & 74.1 & 74.1 & \textbf{74.7} \\
\bottomrule
\end{tabular}
\end{table*}

\noindent \textbf{Effect of Positive-Advantage Suppression.}
We further study step-level positive-advantage suppression (PAS) within IDRL. PAS prevents rejected or ineffective assistant steps from receiving positive reinforcement by setting their positive token advantages to zero, while retaining zero or negative advantages. As shown in Table~\ref{tab:ablation_pas}, removing PAS causes the largest drops on tasks that place greater demands on multi-step reasoning and long-horizon interaction: Search-Oriented performance decreases by 1.2 points, Multimodal Reasoning by 0.9 points, and PinchBench by 0.8 points (77.6 vs.\ 76.8). By contrast, the variant without PAS improves only marginally on Chart Understanding (+0.2) and Real-World Perception (+0.1), where trajectories are typically shorter and intermediate credit assignment is less critical. Overall, PAS improves the average score over 18 benchmarks by approximately 0.4 points (74.7 vs.\ 74.3). These results suggest that trajectory-level rewards alone can incorrectly reinforce ineffective intermediate actions, with the resulting noise becoming more consequential as trajectories grow longer. PAS mitigates this issue by assigning positive credit more selectively, thereby improving learning on reasoning- and interaction-intensive tasks.
\begin{table}[t]
\centering
\caption{Ablation of step-level positive-advantage suppression (PAS) within IDRL. AVG gives equal weight to each of the 18 individual benchmarks. $\Delta$ denotes the difference between IDRL without PAS and IDRL.}
\label{tab:ablation_pas}
\renewcommand{\arraystretch}{1.2}
\begin{adjustbox}{max width=\columnwidth}
\begin{tabular}{lccc}
\toprule
\textbf{Category} & \textbf{IDRL} & \shortstack{\textbf{IDRL} \\ \textbf{w/o PAS}} & \textbf{$\Delta$} \\
\midrule
Chart Understanding   & 82.0 & 82.2 & +0.2 \\
Real-World Perception & 78.4 & 78.5 & +0.1 \\
Multimodal Reasoning  & 79.4 & 78.5 & -0.9 \\
Search-Oriented       & 58.4 & 57.2 & -1.2 \\
Claw-Style            & 77.6 & 76.8 & -0.8 \\
\midrule
\textbf{AVG}          & 74.7 & 74.3 & -0.4 \\
\bottomrule
\end{tabular}
\end{adjustbox}
\end{table}

\subsection{Training Observations and Design Lessons}

\rev{\noindent \textbf{Generator--target model compatibility.}
For SFT data construction, Qwen3.6-27B produced trajectories with a lower raw pass rate than Qwen3.5-397B-A17B, yet training on its filtered data yielded stronger downstream performance. This observation suggests that raw pass rate alone does not determine data quality: a generator that is closer to the target model family may yield a more compatible data distribution, which can be more important than maximizing the generator's standalone success rate.}

\rev{\noindent \textbf{Task-specific interaction budgets.}
We assign different maximum interaction rounds to different task families, using 5 rounds for TIR and 15 rounds for search. A single shared budget can distort behavior: when TIR tasks are given an unnecessarily long horizon, the model may issue uninformative actions, such as generating blank images, merely to consume the available interaction rounds.}

\rev{\noindent \textbf{High-resolution perception for TIR.}
Agentic training on high-resolution images is particularly effective for TIR tasks. On 8K desktop screenshots in HRBench and MME-RealWorld-Lite, allowing the agent to crop and inspect local regions supplies details that may be missed by a single global view, improving fine-grained visual grounding and reasoning.}

\rev{\noindent \textbf{Web-page acquisition for search.}
Search capability depends not only on query quality and retrieval accuracy, but also on extracting detailed evidence from retrieved snippets and web pages. In practice, anti-bot mechanisms can lower the success rate of page-fetching tools. Under RL, repeated failures to retrieve useful information discourage the model from invoking the tool, which can in turn reduce overall search capability. Reliable page acquisition and fine-grained information extraction should therefore be treated as first-class components of the search environment.}

\subsection{Comparison with Business Benchmarks}
\noindent \textbf{Motivation for building a business-specific benchmark.}
Recent multimodal foundation models have achieved strong performance on a wide range of general-purpose benchmarks, demonstrating advances in perception, reasoning, and cross-modal understanding. However, these benchmarks primarily evaluate generic capabilities and do not capture the requirements of real-world content-safety and business-integrity scenarios. Such scenarios have several distinctive properties: (1) policy grounding, where decisions must align with explicit policy provisions; (2) fine-grained semantic discrimination, often involving subtle or borderline cases; (3) multimodal and temporal reasoning or grounding, requiring joint interpretation of visual and textual signals over time; and (4) high-stakes decision making, where errors have direct practical consequences. As a result, performance on existing benchmarks does not reliably translate to effectiveness in these domains.

\noindent \textbf{Pistis Benchmark.}
To address this gap, we construct a large-scale, multimodal, business-specific benchmark, named \textbf{Pistis Benchmark}, that systematically evaluates multimodal models in content-safety and business-integrity scenarios through provision-driven tasks. The benchmark is designed to
(i) cover diverse business scenarios and provision definitions, 
(ii) assess a range of business-specific atomic abilities across modalities and tasks, and 
(iii) support scalable and robust evaluation through approximately 10{,}000 high-quality samples. This framework enables comprehensive comparison between Pistis and competing models, while providing insights for model development and deployment. Pistis Benchmark is compatible with VLMEvalKit~\citep{duan2024vlmevalkit} for standardized evaluation.

\noindent \textbf{Data engineering.} 
To ensure consistency and usability, all collected data are standardized at the video level and aligned with a unified schema.
To improve data quality and diversity, we perform multi-stage multimodal deduplication. At the frame level (intra-video), key frames are selected based on hybrid similarity (low-level features plus deep embeddings), reducing redundancy while preserving representative visual content. At the video level (inter-video), multimodal embeddings built from key frames, OCR, and ASR are used to remove semantically similar videos.
This process removes approximately 25\% of redundant samples, increasing diversity and reducing evaluation bias.

\noindent \textbf{Question-answer generation and filtering.}
We formulate Pistis Benchmark primarily as multimodal VQA tasks, enabling flexible and structured evaluation.
The generation pipeline has two stages. First, an LLM generator produces candidate QA pairs conditioned on the multimodal input and the relevant policy context. Second, an LLM judge filters the candidates by relevance, accuracy, and quality.
The generated questions span multiple formats, including multiple-choice, attribution, action recognition, captioning, reasoning, and grounding tasks. Importantly, questions are designed to require cross-modal reasoning and policy grounding, rather than surface-level recognition.
Finally, we apply additional post-processing to remove low-quality samples and enforce consistency between the QA pair and the original video data.

\noindent \textbf{Benchmark composition and tasks.} Pistis Benchmark comprises approximately 10K high-quality samples spanning both video (71.5\%) and image (28.5\%) inputs, reflecting the predominantly temporal nature of real-world content-safety scenarios. The benchmark is organized into three question formats---multiple-choice, open-ended , and yes/no---and covers a broad spectrum of business-specific atomic abilities. Specifically, reasoning requires policy-grounded inference over multimodal evidence, judging which provision a video may violate and resolving subtle, borderline cases from the interplay between narrative captions and visual composition; VQA covers general question answering over video and image content; and multilingual understanding poses identical policy-relevant questions across languages to ensure consistent judgments. OCR extracts on-screen text verbatim in its original script, while captioning produces neutral, literal descriptions of observable visual content. Action recognition identifies fine-grained physical interactions and attributes, and attribution synthesizes structured product details  into an accurate description. This composition allows Pistis Benchmark to evaluate not only surface-level perception but also fine-grained, policy-grounded, cross-modal decision making.

\noindent \textbf{Evaluation results.}
Table~\ref{tab:exp_results_on_pb} reports results for Pistis and other competitive models on Pistis Benchmark.
The results show that \pahfix{$^\dagger$Pistis-9B (fine-tuned on business-specific data) obtains the highest aggregate score among the compared models (89.7 vs.\ 86.2 for Qwen3.5-9B and 84.3 for Qwen3-VL-8B-Thinking). Pistis improves on OCR, grounding, action recognition, captioning, and video VQA,} which are directly relevant to the content-safety and business-integrity scenarios evaluated by the benchmark.
\begin{table*}[t]
    \scriptsize
    \centering
    \caption{Evaluation of business-specific visual understanding capabilities for content safety and business integrity on Pistis Benchmark. $^\dagger$Pistis-9B is further fine-tuned on business-specific data, whereas all other models are evaluated in their original released form.}
    \label{tab:exp_results_on_pb}
    \resizebox{0.95\linewidth}{!}{
    \begin{tabular}{llcccccc}
    \toprule
    \textbf{Modality} & \textbf{Capability}
    & \makecell{\textbf{InternVL3.5}\\\textbf{8B}\\{\scriptsize thinking}}
    & \makecell{\textbf{Keye-VL-1.5}\\\textbf{8B}\\{\scriptsize thinking}}
    & \makecell{\textbf{Qwen3-VL}\\\textbf{8B}\\{\scriptsize thinking}}
    & \makecell{\textbf{Qwen3.5}\\\textbf{9B}\\{\scriptsize thinking}}
    & \makecell{\textbf{Pistis}\\\textbf{9B}\\{\scriptsize thinking}}
    & \makecell{\textbf{Pistis}$^\dagger$\\\textbf{9B}\\{\scriptsize thinking}} \\
    \midrule
    \multirow{4}{*}{Image}
    & OCR          & 70.8 & 72.2 & 76.6 & 81.4 & 82.1 & 82.8 \\
    & Grounding    & 52.6 & 56.8 & 71.2 & 69.7 & 73.8 & 80.3 \\
    & Attribute    & 84.2 & 96.7 & 99.3 & 99.5 & 99.6 & 99.5 \\
    & Multilingual & 73.9 & 85.9 & 91.5 & 90.7 & 86.5 & 92.3 \\
    \midrule
    \multirow{5}{*}{Video}
    & Reasoning     & 83.1 & 84.3 & 87.3 & 86.0 & 92.5 & 89.5 \\
    & Action recog. & 91.5 & 86.0 & 82.9 & 94.5 & 92.3 & 97.0 \\
    & Caption       & 72.2 & 73.3 & 65.5 & 65.5 & 69.2 & 74.9 \\
    & Multilingual  & 90.4 & 88.1 & 90.9 & 92.2 & 90.5 & 93.3 \\
    & VQA           & 94.7 & 92.8 & 93.1 & 96.2 & 97.7 & 97.8 \\
    \midrule
    \multicolumn{2}{l}{\textbf{Avg. Score}} & 79.3 & 81.8 & 84.3 & 86.2 & 87.1 & \textbf{89.7} \\
    \bottomrule
    \end{tabular}
    }
\end{table*}

\section{Failure Analysis and Future Directions}
\label{sec:failure}

Beyond the aggregate results in Section~\ref{sec:exp}, manual inspection of erroneous Pistis-Agentic rollouts reveals three recurring process-level weaknesses. First, the model may recognize salient visual cues but bind them to an incorrect interpretation of the question. Second, it may use a tool to confirm a preselected hypothesis rather than obtain a decision-relevant measurement. Third, retrieval and reasoning may remain loosely coupled: useful evidence can fail to constrain the final answer, while expressed uncertainty may not trigger retrieval. These failures often arise even when low-level perception is adequate, indicating that stronger benchmark performance does not by itself ensure reliable evidence use.

These patterns help explain the design choices of PAH. The Candidate Ledger preserves explicit candidate--evidence bindings, evidence-driven checkpoints require the model to update this state before answering, and Search Skills provide conditional procedures when retrieval is blocked. However, these mechanisms cannot recover an answer whose correct entity never enters the upstream candidate set. Future work should therefore combine inference-time orchestration with process-level training signals that reward intent verification, decision-relevant tool use, candidate recall, and consistency between retrieved evidence and final answers. Detailed trajectories, visual examples, and a fuller discussion appear in Appendix~\ref{app:failure_cases}.

\section{Conclusion}

We presented the Pistis model family and the general post-training framework behind it. A general recipe, large-scale supervised fine-tuning followed by IDRL, produces two strong specialists that share a reasoning-data SFT foundation: \revdel{Pistis-8B-Thinking} \rev{Pistis-Thinking} for deep multimodal reasoning, and \revdel{Pistis-8B-Agentic} \rev{Pistis-Agentic}, whose SFT stage additionally includes agentic trajectory data before IDRL specialization. Pistis-Agentic shows its strongest gains in multimodal search and also consistently improves Claw-Style interaction at both scales. In particular, Pistis-9B-Agentic exceeds Qwen3.5-9B by 2.8 points on PinchBench, while Pistis-27B-Agentic exceeds Qwen3.6-27B by 0.8 points on the same benchmark. Our ablations further show that IDRL obtains the strongest aggregate Claw-Style score and that removing PAS reduces it, connecting these interaction gains to the proposed training design. More broadly, both Pistis variants perform strongly across multimodal and agentic benchmarks, and interleaving on-policy distillation with reinforcement learning is more stable and effective than performing either alone or jointly optimizing their losses with static weights.

Complementing these model-level contributions, Pistis-Auto-Harnessing (PAH) provides a system-level method for multimodal search: \pahfix{a closed outer loop in which an Optimization Agent proposes, validates, and accepts harness revisions on a development set while the model stays frozen. The resulting Optimized Harness combines a Candidate Ledger, conditionally loaded Search Skills, and an adaptive workflow with budget-aware termination, and improves VDR-testmini accuracy from 26.8\% to 28.6\% under a matched environment interaction budget.} \pahadd{The same frozen Optimized Harness also improves MMSearch, BrowseComp-VL, and LiveVQA without target-set tuning and transfers positively to Seed-2.1-turbo and GPT-5.5. These results broaden the evidence within multimodal search, while transfer to other agent domains such as coding and Claw-Style interaction remains unvalidated.} Together, the model- and system-level results suggest that specialized multimodal capabilities can be cultivated efficiently from a shared foundation, and we hope Pistis serves as a strong basis for future research.

At the same time, the failure analysis in Section~\ref{sec:failure} shows that stronger benchmark performance does not eliminate process-level weaknesses: \revdel{Pistis-8B-Agentic} \rev{Pistis-Agentic} can still mis-bind question intent, use tools to confirm rather than measure, and leave retrieval and reasoning loosely coupled. Closing this gap will require process-level supervision that rewards intent verification, decision-relevant tool use, and consistency between intermediate and final answers, all of which can be integrated into the RL phases of IDRL. We view narrowing the distance between aggregate accuracy and reliable reasoning as a central direction for future work.
\section{Contributors}

\noindent Names within each group are listed alphabetically by surname.

\noindent\textbf{Core Contributors.}
Heyun Chen, Xiaohan Lan, Jiaxi Li, Zhilin Lu, Qi She, Weiwen Xu, Fei Yu, Yujie Zhong.

\noindent\textbf{Contributors.}
Jinghuan Chen, Zijian Feng, Siyu Jiao, Yiheng Lin, Xinhao Wang, Sihan Yang, Jieyu You, Changbin Zhang, Hengyu Zhang, Xudong Zhang, Yunqing Zhao, Shuai Zheng.

\bibliography{colm2024_conference}
\bibliographystyle{colm2024_conference}

\appendix
\clearpage

\section{\texorpdfstring{\pahadd{Additional Analysis of the PAH-Optimized Harness}}{Additional Analysis of the PAH-Optimized Harness}}
\label{app:auto_harness}

This appendix analyzes the frozen Optimized Harness produced by PAH. All statistics below are computed after the harness is frozen and do not feed back into proposal generation or version selection. They characterize the resulting system rather than isolate the causal effect of any single component.

\subsection{Runtime Audit of the Candidate Ledger and Search Skills}

Table~\ref{tab:pah_runtime_audit} summarizes whether the two principal mechanisms actually enter the runtime control flow. The Candidate Ledger is a high-coverage path: 497 of 500 trajectories attempt at least one candidate record, and 490 record one successfully. The first attempt occurs after 2.61 environment interactions on average. By contrast, Search Skills are selectively activated on 135 trajectories, usually after the search has already encountered difficulty.

\begin{center}
\centering
\captionof{table}{Runtime audit of the frozen Optimized Harness on 500 VDR-testmini trajectories.}
\label{tab:pah_runtime_audit}
\begin{adjustbox}{max width=\textwidth}
\begin{tabular}{llr}
\toprule
\textbf{Component} & \textbf{Statistic} & \textbf{Value} \\
\midrule
\multirow{6}{*}{Candidate Ledger} & Trajectories attempting a candidate record & 497/500 (99.4\%) \\
& Trajectories with a successful candidate record & 490/500 (98.0\%) \\
& Successful candidate records per trajectory & 1.522 \\
& Mean interactions before the first record attempt & 2.61 \\
& First attempt immediately after two interactions & 353/497 (71.0\%) \\
& Mean rendered candidate context & 2,455 chars \\
\midrule
\multirow{4}{*}{Search Skills} & Trajectories loading at least one skill & 135/500 (27.0\%) \\
& Successful skill loads & 137 \\
& Loads of \texttt{repair-search-query} & 134/137 (97.8\%) \\
& Mean interactions before the first skill load & 5.51 \\
\bottomrule
\end{tabular}
\end{adjustbox}
\end{center}

The \texttt{repair-search-query} skill is a recovery procedure triggered after a deterministic search failure; it guides the model to diagnose why the previous query was unproductive and formulate a materially revised query before retrieval resumes. No trajectory reaches the ledger context-truncation or update-count cap, indicating that state capacity is not the current bottleneck. The audit instead reveals a useful division of labor: the Candidate Ledger is the routine evidence-state mechanism, whereas Search Skills primarily form a sparse query-recovery path. The broader skill catalog is available but is not yet reliably exercised. These activation statistics are descriptive rather than causal: difficult trajectories are more likely to trigger recovery, so lower accuracy among skill-using trajectories would not imply that the skill itself causes failure.

One possible explanation for the limited use of the broader skill catalog is that these multimodal-search tasks share recurring solution procedures that the policy may already have learned during training, leaving limited room for additional guidance in the form of standard operating procedures (SOPs). However, the present audit does not establish that skills provide little benefit: sparse activation may also reflect limitations of the triggering and routing mechanisms. Controlled skill ablations would be needed to distinguish these explanations and quantify the marginal contribution of skills.

The audit also clarifies the semantics of ledger verification. A \texttt{verified} candidate denotes multi-source support, not proof that visual identity, relation direction, every question premise, and the requested terminal field are all correct. Increasing record frequency or source count alone is therefore not an appropriate optimization objective.

\subsection{A Candidate Ledger Trajectory}

Table~\ref{tab:pah_ledger_case} presents a compact VDR-testmini example in which the query asks how a dress pattern influenced a 1980s horror subgenre and which materials were used for robotic antagonists in a representative film. The ledger does not discard the initial useful style hypothesis when its film hypothesis fails; it preserves the supported portion, exposes the missing material constraint, and allows later evidence to repair the entity.

\begin{table*}[t]
\centering
\caption{Candidate evolution in a successful trajectory. Evidence counts denote source-bound items recorded by the ledger.}
\label{tab:pah_ledger_case}
\begin{adjustbox}{max width=\textwidth}
\begin{tabular}{p{0.15\textwidth}p{0.28\textwidth}p{0.36\textwidth}p{0.12\textwidth}}
\toprule
\textbf{Stage} & \textbf{Candidate state} & \textbf{Evidence gap and update} & \textbf{Role} \\
\midrule
Initial hypothesis & Comic-book/pop-art style; \emph{X-Tro} as a tentative film & Two items support the style direction, but the film cannot satisfy the robotic-material constraint. & Preserve style; reject entity \\
Entity repair & Switch to \emph{Chopping Mall}; record fiberglass, foam, and supporting production details & Three items establish the robotic antagonists and their construction, while the style-to-subgenre relation remains incomplete. & Repair entity; fill field \\
Evidence closure & Graphic, action-oriented techno-/sci-fi horror; robots primarily made from fiberglass and foam & Six items jointly cover the style, subgenre, film identity, antagonists, and requested materials without direct contradiction. & Support final answer \\
\bottomrule
\end{tabular}
\end{adjustbox}
\end{table*}

This trajectory illustrates the intended role of the ledger: it makes hypothesis revision and constraint coverage explicit while leaving every evidence action and the final answer to the same frozen reasoning model.

\subsection{Transfer across Frozen Policy Models}
\label{app:pah_policy_transfer}

The Pistis-27B-Agentic row in Table~\ref{tab:pah_policy_transfer} reproduces the primary matched-budget comparison in Table~\ref{tab:auto_harness_results}. We then retain the same frozen Optimized Harness while replacing the reasoning policy with Seed-2.1-turbo and GPT-5.5. No harness revision or model-specific version selection is performed. All three policies improve relative to their corresponding Baseline Harness, while their average environment interactions remain close to the matched baselines.

\begin{center}
\centering
\captionof{table}{Transfer of the same frozen Optimized Harness across reasoning policies on VDR-testmini.}
\label{tab:pah_policy_transfer}
\begin{adjustbox}{max width=\textwidth}
\begin{tabular}{lrrrr}
\toprule
\textbf{Frozen policy} & \textbf{Baseline acc.} & \textbf{Optimized acc.} & \textbf{Gain} & \textbf{Avg. interactions (base/opt.)} \\
\midrule
Pistis-27B-Agentic & 26.8 & 28.6 & +1.8 & 8.68 / 8.60 \\
Seed-2.1-turbo & 27.6 & 29.0 & +1.4 & 7.31 / 7.69 \\
GPT-5.5 & 28.6 & 32.6 & +4.0 & 6.44 / 6.49 \\
\bottomrule
\end{tabular}
\end{adjustbox}
\end{center}

For GPT-5.5, the paired comparison contains 47 positive and 27 negative flips among 74 discordant examples. Its larger gain suggests that stronger reasoning policies may use structured state and conditional procedures more effectively, but the present experiments do not isolate this interaction causally.

\subsection{Design Lessons and Limitations}

\noindent\textbf{Design lessons.}
The optimization history yields four practical lessons. First, mechanism activation is only a diagnostic: a revision is accepted only when the complete development-set task metric improves. Second, guidance should be conditionally exposed from replayable states; globally persistent prompts can perturb trajectories that were already correct. Third, upstream identity and relation errors should be addressed before strengthening downstream field extraction, since stricter completion around a wrong entity can reinforce the wrong answer. Fourth, each candidate version should be independently derived from the current best harness and remain fully reversible, so rejected mechanisms do not silently accumulate.

\noindent\textbf{Limitations.}
The frozen Optimized Harness can still over-commit to an early incumbent because it lacks a systematic evidence-backed challenger test. Apparent source diversity can also be overstated when several evidence identifiers derive from the same underlying material. Relation endpoints and the direct binding between the input image and a textual entity remain incompletely certified. Finally, budget-aware termination can occasionally prevent one last decisive verification, while most skills other than query repair are not yet reliably activated. These limitations concern the current Optimized Harness; whether PAH discovers different and stronger mechanisms for other agent domains, such as coding or Claw-Style interaction, requires separate development/test studies.
\clearpage
\section{Detailed Failure Cases}
\label{app:failure_cases}

This appendix expands the process-level failure analysis summarized in Section~\ref{sec:failure}. We present representative cases from multimodal reasoning, tool-integrated reasoning, and agentic search. Across these settings, low-level perception is often adequate; the central weakness is how the reasoning process interprets the task, gathers decision-relevant evidence, and binds that evidence to the final answer.

\subsection{Multimodal Reasoning: Salient Cues Are Read, Intent Is Not}

\begin{figure}[t]
    \centering
    \includegraphics[width=\linewidth]{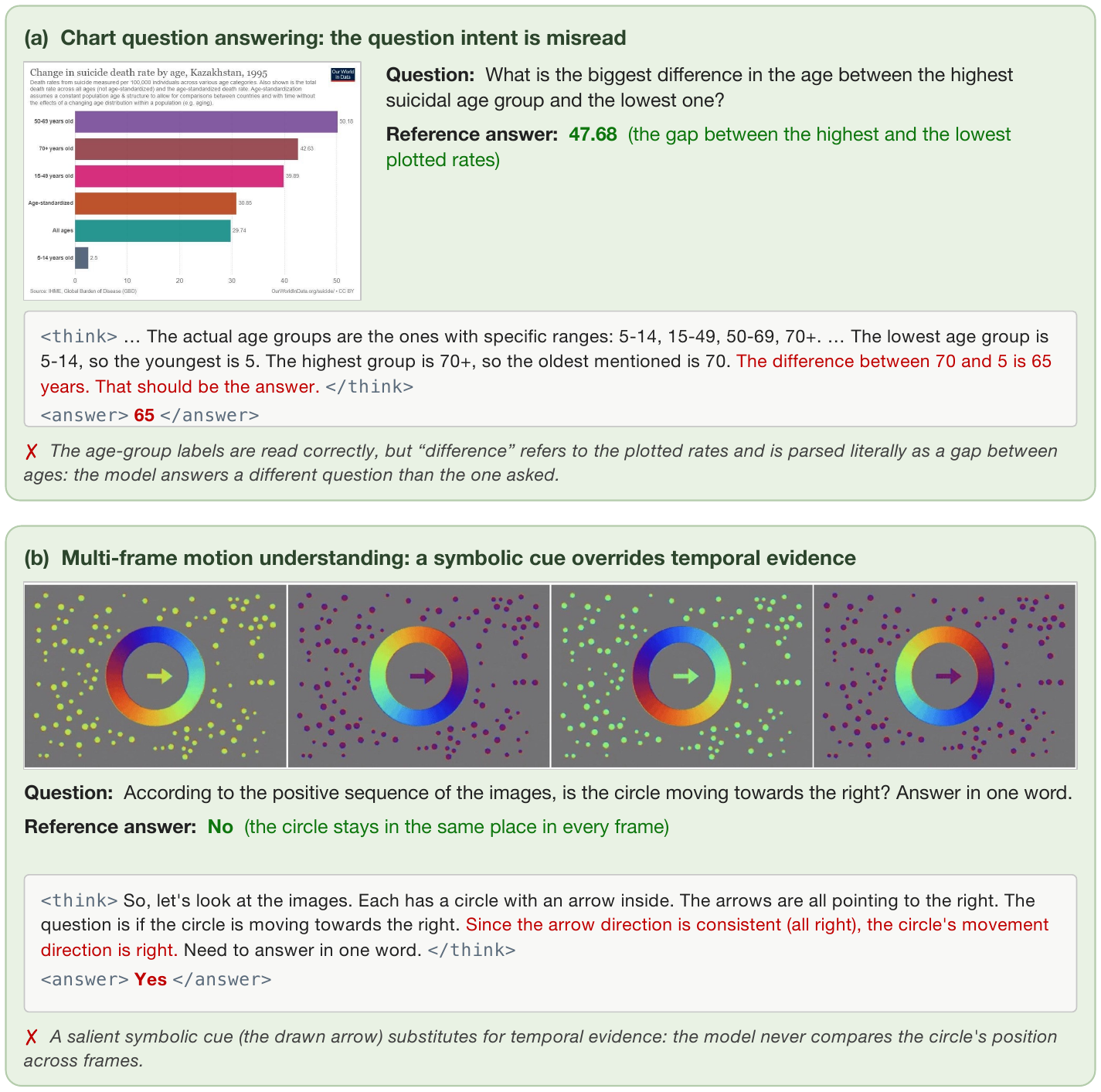}
    \caption{Representative failure cases of \pahfix{Pistis-Agentic} in multimodal reasoning. (a) The age-group labels are read correctly, but the question intent (the gap between the highest and lowest plotted rates) is parsed literally as a gap between ages. (b) The drawn arrow, a salient symbolic cue, overrides the cross-frame evidence that the circle never moves. Model excerpts are abridged; red marks the faulty steps and final answers, and green marks the reference answers.}
    \label{fig:failure_reasoning}
\end{figure}

In Figure~\ref{fig:failure_reasoning}(a), the model identifies the age-group labels correctly but interprets ``difference in the age'' literally as the gap between group ages, rather than the intended gap between the plotted rates, and therefore answers a different question. In Figure~\ref{fig:failure_reasoning}(b), when asked whether the circle moves to the right across an ordered sequence of frames, the model latches onto the right-pointing arrow drawn inside the circle and never compares the circle's position across frames. Neither failure is due to missing the salient visual elements: in both cases, the model anchors on the most literal cue instead of verifying its interpretation against the visual evidence.

\subsection{Tool-Integrated Reasoning: Tools Confirm Instead of Measure}

\begin{figure}[t]
    \centering
    \includegraphics[width=\linewidth]{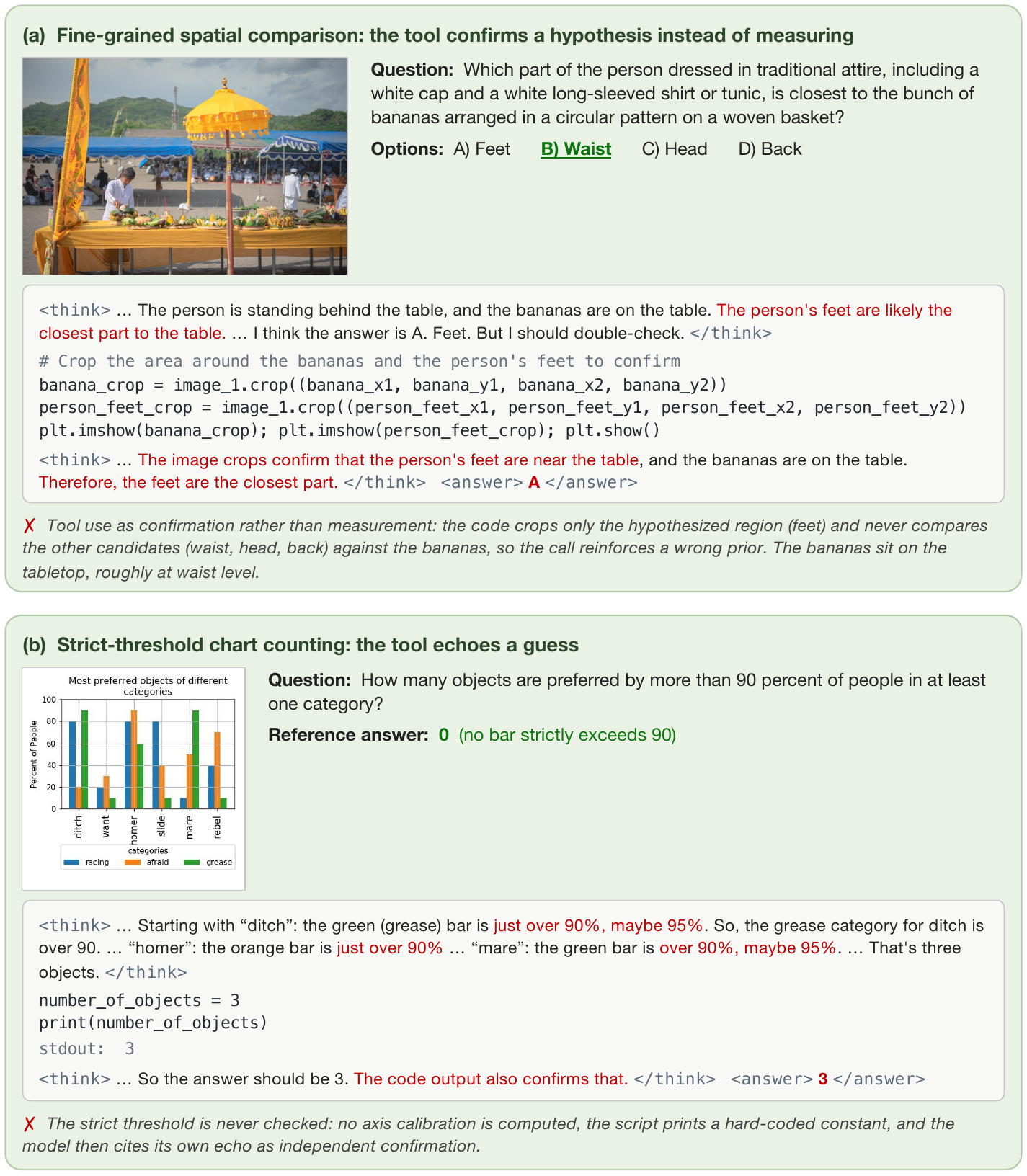}
    \caption{Representative failure cases of Pistis-Agentic in tool-integrated reasoning. In both rollouts, the code interpreter is used to confirm a preselected hypothesis rather than to measure: (a) only the hypothesized region is cropped and the candidate comparison is never performed; (b) the script prints a hard-coded count, and the printed output is then cited as confirmation. Excerpts are abridged; red marks the faulty steps.}
    \label{fig:failure_tir}
\end{figure}

Figure~\ref{fig:failure_tir} shows two rollouts with a code interpreter. In (a), the model first forms the hypothesis that the feet are closest to the target object, then writes code that crops only the target and the presumed feet region; the waist, head, and back are never examined, so the visualization merely reinforces the prior. In (b), the question requires a strict greater-than-90 comparison; the model eyeballs the bars as ``just over 90'', writes a script that prints a hard-coded count, and then cites the printed output as confirmation. In both cases, the tool call is formulated to support a preselected answer rather than produce decision-relevant measurements, and an executed cell that adds no information is treated as independent evidence.

\subsection{Agentic Search: Retrieval and Reasoning Are Loosely Coupled}

\begin{figure}[t]
    \centering
    \includegraphics[width=\linewidth]{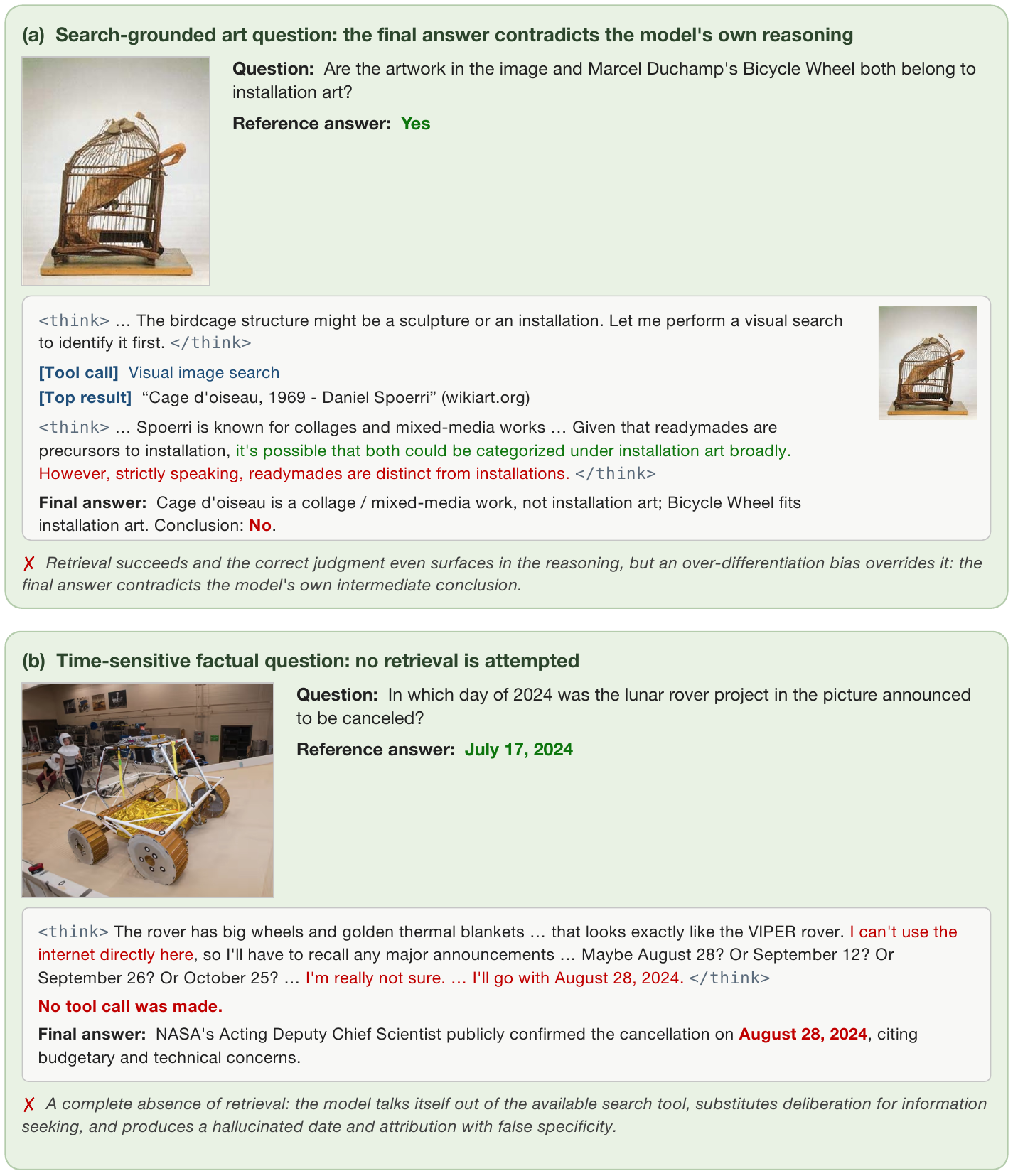}
    \caption{Representative failure cases of Pistis-Agentic in agentic search. (a) Retrieval succeeds and the correct judgment (green) surfaces in the reasoning, yet the final answer contradicts it. (b) The model never invokes the available search tool and produces a hallucinated date with fabricated attribution. Excerpts are abridged; red marks the faulty steps.}
    \label{fig:failure_agentic}
\end{figure}

Figure~\ref{fig:failure_agentic} illustrates two complementary failures of the search--reasoning loop. In (a), visual search succeeds and the correct judgment even surfaces in the reasoning trace, yet the final answer contradicts it: the model over-differentiates between closely related concepts and overrides its own intermediate conclusion, so the retrieved evidence does not bind the final answer. In (b), the converse occurs: facing a time-sensitive factual question, the model talks itself out of using the available search tool, enumerates candidate dates from memory while explicitly acknowledging uncertainty, and finally emits a hallucinated date with fabricated attribution. The uncertainty is verbalized but never operationalized into a tool call.

These observations help explain, rather than retrospectively motivate, the PAH design in Section~\ref{sec:auto_harness}. Its Candidate Ledger keeps retrieved facts bound to explicit candidates with traceable sources, its checkpoints ask the model to update this state before answering, and its Search Skills supply operating procedures when retrieval is blocked. The same audit also exposes a remaining limitation: if the correct entity never enters the upstream candidate set, downstream evidence organization cannot recover it.

\subsection{Implications for Future Work}

The cases suggest three complementary training directions. First, \emph{intent grounding} should teach the model to state and verify its interpretation of an underspecified question against visual evidence before committing to a computation. Second, \emph{measurement-grounded tool use} should reward calls that generate decision-relevant evidence, such as comparing all candidate regions or calibrating chart axes before a threshold judgment, while penalizing confirmatory no-op calls. Third, \emph{retrieval--reasoning coupling} should reward candidate recall, consistency between intermediate conclusions and final answers, and the conversion of expressed uncertainty into targeted retrieval. These rule-checkable trajectory signals can be incorporated into the RL phases of IDRL, while its distillation phases transfer the corresponding behaviors from a stronger teacher. We leave systematic quantitative evaluation of these failure modes to future work.

\end{document}